# Machine Learning for Microcontroller-Class Hardware: A Review

Swapnil Sayan Saha⦿, *Graduate Student Member, IEEE*, Sandeep Singh Sandha, and Mani Srivastava, *Fellow, IEEE*

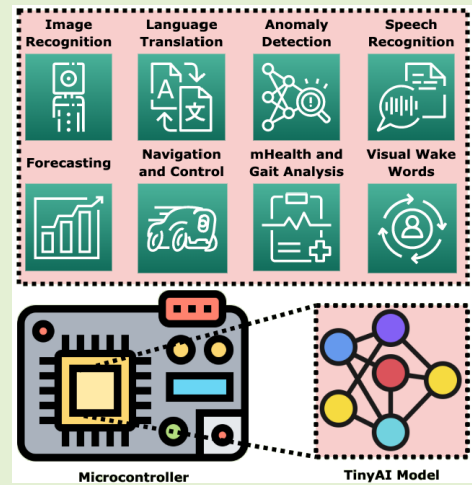

*Abstract*—The advancements in machine learning (ML) opened a new opportunity to bring intelligence to the low-end Internet-of-Things (IoT) nodes, such as microcontrollers. Conventional ML deployment has high memory and computes footprint hindering their direct deployment on ultraresource-constrained microcontrollers. This article highlights the unique requirements of enabling onboard ML for microcontroller-class devices. Researchers use a specialized model development workflow for resource-limited applications to ensure that the compute and latency budget is within the device limits while still maintaining the desired performance. We characterize a closed-loop widely applicable workflow of ML model development for microcontroller-class devices and show that several classes of applications adopt a specific instance of it. We present both qualitative and numerical insights into different stages of model development by showcasing several use cases. Finally, we identify the open research challenges and unsolved questions demanding careful considerations moving forward.

## I. INTRODUCTION

LOW-END Internet-of-Things (IoT) nodes, such as microcontrollers, are widely adopted in resource-limited applications, such as wildlife monitoring, oceanic health tracking, search and rescue, activity tracking, industrial machinery debugging, onboard navigation, and aerial robotics [1], [2]. These applications limit the compute device payload capabilities and necessitate the deployment of lightweight hardware and inference pipelines. Traditionally, microcontrollers operated on low-dimensional structured sensor data (e.g., temperature and humidity) using classical methods, making simple inferences at the edge. Recently, with the advent of machine learning (ML), considerable endeavors are underway to bring ML to the edge [3], [4].

However, directly porting ML models designed for high-end edge devices, such as mobile phones or single-board computers, are not suitable for microcontrollers. A typical microcontroller has 128-kB RAM and 1 MB of flash, while a mobile phone can have 4 GB of RAM and 64 GB of storage [5]. The ultraresource limitations of microcontroller-class IoT nodes demand the design of a systematic workflow and tools to guide onboard deployment of ML pipelines.

This article presents the unique requirements, challenges, and opportunities presented when developing ML models doing sensor information processing on microcontrollers. While prior surveys [3], [4], [6], [7] present a qualitative review of the model development cycle for microcontrollers, they fail to provide quantitative comparisons across alternative workflow choices and insights from application-specific case studies. In contrast, we illustrate a closed-

Manuscript received 5 June 2022; revised 18 July 2022; accepted 27 September 2022. Date of publication 5 October 2022; date of current version 14 November 2022. This work was supported in part by the CONIX Research Center, one of six centers in Joint University Microelectronics Program (JUMP), a Semiconductor Research Corporation (SRC) Program sponsored by the Defense Advanced Research Projects Agency (DARPA); in part by the IoBT REIGN Collaborative Research Alliance funded by the Army Research Laboratory (ARL) under Cooperative Agreement W911NF-17-2-0196; and in part by the NIH mHealth Center for Discovery, Optimization and Translation of Temporally-Precise Interventions (mDOT) under Award 1P41EB028242. The associate editor coordinating the review of this article and approving it for publication was Dr. Ashish Pandharipande. *(Corresponding author: Swapnil Sayan Saha.)*

Swapnil Sayan Saha and Mani Srivastava are with the Department of Electrical and Computer Engineering and the Department of Computer Science, University of California at Los Angeles, Los Angeles, CA 90095 USA (e-mail: swapnilsayan@g.ucla.edu; mbs@ucla.edu).

Sandeep Singh Sandha is with Amazon, Seattle, WA 98121 USA (e-mail: ssandha@ucla.edu).

Digital Object Identifier 10.1109/JSEN.2022.3210773

TABLE I
COMPARISON OF HARDWARE FOR DOING ML ON CLOUD SERVERS,
MOBILE PHONES, AND MICROCONTROLLERS [8]

| Platform | Memory | Storage | Power |
| --- | --- | --- | --- |
| Cloud GPU | 16 GB HBM | TB/PB | 250W |
| Mobile CPU | 4 GB DRAM | 64 GB Flash | 8W |
| Microcontroller | 2-1024 kB SRAM | 32-2048 kB eFlash | 0.1-0.3W |

loop workflow of ML model development and deployment for microcontroller-class IoT nodes with quantitative evaluation, numerical analysis, and benchmarks showing different instances of proposed workflow across various applications. Specifically, we discuss, in detail, workflow components while making performance comparisons and tradeoffs of the workflow adoptions in the existing literature. Finally, we also identify bottlenecks in the current model development cycle and propose open research challenges going forward. Our contributions are given as follows.

1) We illustrate a coherent and closed-loop ML model development and deployment workflow for microcontrollers. We delineate each block in the workflow, providing both qualitative and numerical insights.
2) We provide application-dependent quantitative evaluation and comparison of proposed workflow adaptations.
3) We discuss several tradeoffs in the existing model-development process for microcontrollers and showcase opportunities and ideas in this workspace.

The rest of this article is organized as follows. Section II outlines the TinyML workflow of model development and deployment for microcontrollers. Section III explores data engineering frameworks. Section IV shows feature projection techniques. Section V discusses model compression methods. Section VI describes lightweight ML blocks suitable for microcontrollers. Section VII discusses neural architecture search (NAS) frameworks for microcontrollers. Section VIII outlines several software suites available for porting developed models onto microcontrollers. Section IX showcases TinyML online learning frameworks. Section X provides a quantitative and qualitative comparison of workflow variations depending on the application. Section XI presents interrelative and quantitative analysis of individual portions of the workflow through case studies. Section XII illustrates open challenges and ideas for future research. Section XIII provides concluding remarks.

## II. TINYML WORKFLOW

We use the term 'TinyML' to refer to model compression, machine-learning blocks, AutoML frameworks, and hardware and software suites designed to perform ultralow-power ($\leq 1$ mW), always-on, and onboard sensor data analytics [4], [6], [7] on resource-constrained platforms. Typical TinyML platforms, such as microcontrollers, have SRAM in the order of $10^0$–$10^2$ kB and flash in the order of $10^3$ kB [6]. Table I provides characteristics of these devices compared to cloud servers and mobile phones. Given the widespread penetration of microcontroller-based IoT platforms in our daily lives for pervasive perception-processing-feedback applications, there is a growing push toward embedding intelligence into these frugal smart objects [3]. Embedded AI on microcontrollers is motivated by *applicability*, *independence from network infrastructure*, *security and privacy*, and *low deployment cost*.

### A. Applicability

Neural networks have been shown to provide rich and complex inferences over the first-principle approaches for sensor data analytics without domain expertise. With the emergence of real-time ML for microcontrollers, it is possible to turn IoT nodes from simple data harvesters or first-principles data processors to learning-enabled inference generators. TinyML combines the lightweightness of first-principle approaches with the accuracy of large neural networks.

### B. Independence From Network Infrastructure via Remote Deployment

Traditionally, sensor data are offloaded onto models running on mobile devices or cloud servers [19], [20]. This is not suitable for time-critical sense-compute-actuation applications, such as autonomous driving [21], [22], robot control [4], [23], and industrial control system. Moreover, reliable network bandwidth or power may not be available for communicating with online models, such as in wildlife monitoring [1] or energy-harvesting intermittent systems [24], [25], [26]. TinyML allows offline and onboard inference without requiring data offloading or cloud-based inference.

### C. Security and Privacy

Streaming private data onto third-party cloud servers yields privacy concerns from end-users, while cybercriminals can exploit weakly protected data streams. Federated learning (FL) [27], secure aggregation [28], and homomorphic encryption [29] allow privacy-preserving and secure inference but suffer from expensive network and compute requirement. Onboard inference constrains the source and destination of private data within the IoT node itself, reducing the probability of privacy leaks and attack surfaces.

### D. Low Deployment Cost

While graphics processing units (GPUs) have revolutionized deep-learning [30], GPUs are energy-hungry and expensive to maintain continually for inference using small models, leading to long-term financial and environmental degeneration [5]. A Cortex M4 class microcontroller costs around 5–10 USD and can run on a coin-cell battery for months, if not years [7]. TinyML allows these microcontrollers to be exploited for ultralow-power and low-cost AI inference.

Achieving *low deployment cost* without sacrificing *performance gains* requires a unique workflow to port ML models onto microcontrollers compared to traditional model design. Fig. 1 illustrates the general "closed-loop" workflow for TinyML model development and deployment. For various parts of this workflow, specific technologies and variations have emerged [6], [8], [31], which we discuss in upcoming sections. The workflow can be divided into two phases.

TABLE II
MLPerf Tiny v0.5 Inference Benchmarks [9]

| Application | Dataset (Input Size) | Model Type (TFLM model size) | Quality Target (Metric) |
|---|---|---|---|
| Keyword Spotting | Speech Commands [10] (49×10) | DS-CNN [11] [12] [13] (52.5 kB) | 90% (Top-1) |
| Visual Wake Words | VWW Dataset [14] (96×96) | MobileNetV1 [12] (325 kB) | 80% (Top-1) |
| Image Recognition | CIFAR-10 [15] (32×32) | ResNetV1 [16] (96 kB) | 85% (Top-1) |
| Anomaly Detection | ToyADMOS [17], MIMII [18] (5×128) | FC-Autoencoder [9] (270 kB) | 0.85 AUC |

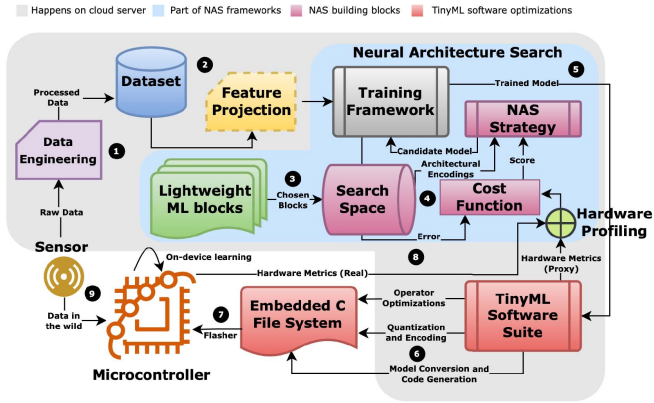

Fig. 1. Closed-loop workflow of porting ML models onto microcontrollers. Step (3)–Step (8) are repeated until the desired performance is achieved. (1) Data engineering performs acquisition, analytics, and storage of raw sensor streams (see Section III). (2) Optional feature projection directly reduces dimensionality of input data (see Section IV). (3) Models are chosen from a lightweight ML zoo based on the application and hardware specifications (see Sections VI and X). (4) NAS strategy builds candidate models from the search space for training and evaluates the model based on cost function (see Section VII). (5) Trained candidate model is ported to a TinyML software suite. (6) TinyML software suite performs inference engine optimizations, deep compression, and code generation. It also provides approximate hardware metrics (e.g., SRAM, flash, and latency) (see Sections V, VII, and VIII). (7) Embedded C file system is ported onto the microcontroller via command line interface. (8) Microcontroller optionally reports real runtime hardware metrics back to the NAS strategy (see Section VII). (9) On-device training or FL is used occasionally to account for shifts in incoming data distribution (see Section IX).

*1) Model Development Phase:* The phase begins by preparing a dataset from raw sensor streams using **data engineering** techniques (see Section III). Data engineering frameworks are used to collect, analyze, label, and clean sensory streams to produce a dataset. Optionally, **feature projection** (see Section IV) is also performed at this stage. Feature projection reduces the dimensionality of the input data through linear methods, nonlinear methods, or domain-specific feature extraction. Next, several models are chosen from a pool of established **lightweight model zoo** based on the application and hardware constraints (see Sections VI and X). The zoo contains optimized blocks for well-known machine-learning primitives (e.g., convolutional neural networks, recurrent neural networks (RNNs), decision trees (DTs), k-nearest neighbors (kNNs), convolutional-recurrent architectures, and attention mechanisms). To achieve maximal accuracy within microcontroller SRAM, flash, and latency targets, **NAS** or hyperparameter tuning is performed on candidate models from the zoo (see Section VII). The hardware metrics are either obtained through proxies (approximations) or real measurements.

*2) Model Deployment Phase:* The deployment phase begins by porting the best performing model to a **TinyML software suite** (see Section VIII). These suites perform inference engine optimizations, operator optimizations, and **model compression** (see Section V), along with embedded code generation. The embedded C file system is then flashed onto the microcontroller for inference. The model can be periodically fine-tuned to account for data distribution shifts using **online learning** (on-device training and FL) frameworks (see Section IX).

To measure and compare the performance of the tinyML workflow for specific applications, Banbury et al. [9] proposed the widely used MLPerf Tiny Benchmark Suite, as illustrated in Table II. The benchmark contains four tasks representing a wider array of applications expected from microcontroller-class models. These include multiclass image recognition, binary image recognition, keyword spotting, and outlier detection. The benchmark suite also embraces the usage of standard datasets for each task and provides quality target metrics and model size that new workflows should aim to achieve. Hardware metrics include the working memory requirements (SRAM), model size (flash), number of multiply and add operations (MACs), and latency. From Sections III to IX, we discuss each block in the TinyML workflow, while, in Section X, we provide quantitative evaluation of the entire workflow based on applications in light of the benchmarks. In Section XI, we break down the end-to-end workflow and provide an analysis of individual aspects.

## III. Data Engineering

Data engineering is the practice of building systems for acquisition, analytics, and storage of data at scale [37]. Data engineering is well explored in production-scale big data systems, where robust and scalable analytics engines (e.g., Apache Spark, Apache Hadoop, Apache Hive, Apache H2O, Apache Flink, and DataBricks LakeHouse) ingest real-time sensory data via publish-subscribe paradigms (e.g., MQTT and Apache Kafka) [38]. Data streaming systems provide real-time data acquisition protocols for requirement definitions and data gathering, while analytics engines provide support for data provenance, refinement, and sustainment. Popular general-purpose exploratory data analysis tools used in TinyML data analytics include MATLAB [39], Giotto-TDA [40], OpenCV [41], ImgAug [42], Pillow [43], Scikit-learn [44], and SciPy [45]. To suit the specific needs and goals of data engineering for TinyML systems, several specialized frameworks have emerged, as illustrated in Table III.

A major challenge for enabling applications that use ML on microcontrollers is preparing the data and learning techniques that can automatically generalize well on unseen scenarios [33]. Thereby, most of these frameworks provide common data augmentation and data cleaning techniques, such as geometric transforms, spectral transforms, oversampling, class balancing, and noise addition. MSWC [33] and Plumerai Data [36] go one step further, providing unit tests and anomaly detectors to identify problematic samples and evaluate the quality of labeled data. Plumerai Data can also automatically identify samples in the training set that is likely to be edge cases or problematic based on model performance on detected problematic samples. Such test-driven development can help users discover edge cases and outliers during model validation stages, and allow users to apply targeted augmentation, oversampling, and label correction. To reduce data collection bias, labeling errors, and manual labeling effort, Edge Impulse [32], MSWC [33], SensiML DCL [34], and Plumerai Data [36] provide AI, DSP, and heuristic-assisted automated labeling tools. In particular, for large-scale keyword spotting dataset generation, MSWC can automatically estimate word boundaries from audio with transcription using forced alignment and extract keywords based on user-defined heuristics in 50 languages. MSWC also automatically ensures that the generated dataset is balanced by gender and speaker diversity. Edge Impulse provides automated labeling of object detection data using YOLOv5 and extraction of word boundaries from keyword spotting audio samples using DSP techniques. SensiML DCL allows video-assisted threshold-based semiautomated labeling of sensor data. Overall, these frameworks ensure that the data being used for training are relevant in context, free from bias, class-balanced, correctly labeled, contains edge cases, free from shortcuts, and encompass sufficient diversity [36].

TABLE III
FEATURES OF NOTABLE TINYML DATA ENGINEERING FRAMEWORKS

| Framework | Data Type | Collection | Labeling | Alignment | Augmentation | Visualization | Cleaning | Open-source |
|---|---|---|---|---|---|---|---|---|
| Edge Impulse [32] | Audio, images, time-series | Real-time (WebUSB, serial daemon, Linux SDK), offline | AI-assisted, DSP-assisted, manual | ✗ | Geometric image transforms, noise, audio spectrogram transforms, color depth | Images, plots: raw, spectrogram, statistical, DSP, MFE, MFCC, syntiant, feature explorer | Class balancing, crop, scale, split | ✗ |
| MSWC [33] | Audio with transcription | Offline speech datasets | Heuristic-based auto | Montreal forced | Synthetic noise, environmental noise | Plots: raw, spectrogram, feature embeddings | Gender balance, speaker diversity, self-supervised quality estimation | ✓ |
| SensiML DCL [34] | Time-series, audio | Real-time (WiFi, BLE, Serial daemon), offline | Plot-assisted, Threshold-based auto | Video-assisted | Noise, pool, convolve, drift, dropout, quantize, reverse, time warp | Plots: raw, spectrogram, statistical, DSP, MFCC | Class balancing, crop, scale, split | ✗ |
| Qeexo AutoML [35] | Time-series, audio | Real-time (Serial daemon, BLE), offline | Plot-assisted | ✗ | ✗ | Plots: raw, spectrogram, statistical, DSP, MFCC, feature embeddings | Segment | ✗ |
| Plumerai Data [36] | Images | Offline | AI-assisted, manual | ✗ | Targetted image transforms, oversampling | Images (AI-assisted visual similarity) | Unit tests, failure case identification | ✗ |

## IV. FEATURE PROJECTION

An optional step in the TinyML workflow is to directly reduce the dimensionality of the data. Models operating on intrinsic dimensions of the data are computationally tractable and mitigate the curse of dimensionality. Feature projection can be divided into three types.

### A. Linear Methods

Linear methods for dimensionality reduction commonly used in large-scale data mining include matrix factorization and principal component analysis (PCA) techniques, such as singular value decomposition (SVD) [61], flattened convolutions [61], non-negative matrix factorization (NMF) [62], independent component analysis (ICA) [63], and linear discriminant analysis [64]. PCA is used to maximize the preservation of variance of the data in the low-dimensional manifold [65]. Among the popular linear methods, NMF is suitable for finding sparse, part-based, and interpretable representations of nonnegative data [62]. SVD is useful for finding a holistic yet deterministic representation of input data with a hierarchical and geometric basis ordered by correlation among the most relevant variables. SVD provides a deeper factorization with lower information loss than NMF. ICA is suitable for finding independent features (blind source separation) from non-Gaussian input data [63]. ICA does not maximize variance or mutual orthogonality among the selected features. Nevertheless, linear methods are unable to model nonlinearities or preserve the global relationship among features and struggle in presence of outliers, skewed data distribution, and one-hot encoded variables.

### B. Nonlinear Methods

Nonlinear methods minimize a distance metric (e.g., fuzzy embedding topology [66], Kullback–Leibler divergence [67], local neighborhoods [68], and Euclidean norm [69]) between the high-dimensional data and a low-dimensional latent representation. Nonlinear methods to handle nonlinear sampling of low-dimensional manifolds by high-dimensional vectors include locally linear embedding (LLE) [68], kernel PCA [69], t-distributed stochastic neighbor embedding (t-SNE) [70], uniform manifold approximation and projection (UMAP) [66], and autoencoders [71]. Kernel PCA couples k-NN, Dijkstra's algorithm, and partial eigenvalue decomposition to maintain geodesic distance in a low-dimensional space [69]. Similarly, LLE can be thought of as a PCA ensemble maintaining local neighborhoods in the embedding space, decomposing the latent space into several small linear functions [68]. However, both LLE and kernel PCA do not perform well with large and complex datasets. t-SNE optimizes KL-divergence between student's T distribution in the manifold-space and Gaussian joint probabilities in the higher dimensional space [70].

t-SNE is able to reveal data structures at multiple scales, manifolds, and clusters. Unfortunately, t-SNE is computationally expensive, lacks explicit global structure preservation, and relies on random seeds. UMAP optimizes a low-dimensional fuzzy embedding to be as topologically similar as the Cech complex embedding [66]. Compared to t-SNE, UMAP provides a more accurate global structure representation while also being faster due to the use of graph approximations. Nonetheless, while linear methods have been ported to microcontrollers [32], [72], nonlinear methods are not suitable for real-time execution on microcontrollers and are usually used for visualizing high-dimensional handcrafted features.

### C. Feature Engineering

Feature engineering uses domain expertise to extract tractable features from the raw data [73]. Typical features include spectral and statistical features. Domain-specific feature extraction is generally more suited for microcontrollers over linear and nonlinear dimensionality reduction techniques due to their relative lightweightness, as well as the availability of dedicated signal processors in commodity microcontrollers for spectral processing. However, feature engineering requires human knowledge to design statistically significant features. Feature selection can reduce the number of redundant features further during model development [74]. Feature selection methods include statistical tests, correlation modeling, information-theoretic techniques, tree ensembles, and metaheuristic methods (e.g., wrappers, filters, and embedded techniques) [75].

## V. PRUNING, QUANTIZATION, AND ENCODING

Model compression aims to reduce the bitwidth and exploit the redundancy and sparsity inherent in neural networks to reduce memory and latency. Han et al. [49] first showed the concept of pruning, quantization, and Huffman coding jointly in the context of pretrained deep neural networks (DNNs). Pruning [76] refers to masking redundant weights (i.e., weights lying within a certain activation interval) and representing them in a row form. The network is then retrained to update the weights for other connections. Quantization [77] accelerates DNN inference latency by rounding off weights to reduce bit width while clustering similar ones for weight sharing. Encoding (e.g., Huffman encoding) represents common weights with fewer bits, either through conversion of dense matrices to sparse matrices [49] or smaller dense matrices through parameter redundancies [78]. Combining the three techniques can drastically reduce the size of state-of-the-art DNNs, such as AlexNet (35×, 6.9 MB), LeNet-5 (39×, 44 kB), LeNet-300-100 (40×, 27 kB), and VGG-16 (49×, 11.3 MB) without losing accuracy [49].

### A. Common Model Compression Techniques

Table IV showcases and compares several frameworks for model compression for microcontrollers. Among the different frameworks, TensorFlow Lite [46] is available as part of the TensorFlow training framework [79], while others are standalone libraries that can be integrated with TensorFlow or PyTorch. 88% of the frameworks provide various quantization primitives, while 50% of the frameworks support several pruning algorithms. Most of these techniques result in unstructured or random sparse patterns.

*1) Quantization Schemes:* From Table IV, we can observe that the most widely used quantization technique for microcontrollers is the fixed-precision uniform affine **posttraining quantizer**, where a real number is mapped to a fixed-point representation via a scale factor and zero-point (offset) after training [47], [80]. Variations include quantization of weights, weights, and activations, and weights, activations, and inputs [81]. While posttraining quantization (with 4, 8, and 16 bits) has been shown to reduce the model size by 4× and speed up inference by 2–3×, **quantization-aware training** is recommended for microcontroller-class models to mitigate layerwise quantization error due to a large range of weights across channels [47], [80]. This is achieved through the injection of simulated quantization operations, weight clamping, and fusion of special layers [51], allowing up to 8× model size reduction for same or lower accuracy drop. However, care must be taken to ensure that the target hardware supports the used bitwidth. To account for distinct compute and memory requirements of different layers, **mixed-precision quantization** assigns different bitwidths for weights and activation for each layer [82]. For microcontrollers, the network subgraph is represented as a quantized convolutional layer with vectorized MAC unit, while special layers are folded into the activation function via integer channel normalization [55], [83]. Mixed-precision quantization provides 7× memory reduction [55] but is supported by limited models of microcontrollers. Recently, **binarized neural networks** [84] have been ported onto microcontroller-class hardware [52], where the weights and activations are quantized to a single bit ($-1$ or $+1$). Binarized quantization can provide 8.5–19× speedup and 8× memory reduction [53].

*2) Pruning Algorithms:* Among the different pruning algorithms, **weight pruning** is the most common, providing 4× speedup and 5–10× memory reduction [48], [49]. Weight pruning follows a schedule that specifies the type of layers to consider, the sparsity distribution to follow during training or fine-tuning, and the metric to follow when pruning (pruning policy). Common weight pruning evaluation metrics include the level and norm of weights [54], [79]. For intermittent computing systems with extremely limited power budgets, the pruning policy usually includes the energy and memory budget to maximize the collection of interesting events per unit of energy [24]. Pruning policies for intermittent computing treat pruning as a hyperparameter tuning problem, sweeping through the memory, energy, and accuracy spaces to build a Pareto frontier. Some frameworks [51], [54] provide support for structured pruning, allowing policies for channel and filter pruning rather than pruning weights in an irregular fashion.

### B. Structured Sparsity

Although model compression improves speedup, eliminates ineffective computations, and reduces storage and memory access costs, unstructured sparsity can induce irregular processing and waste execution time. The benefits of efficient acceleration through sparsity require special hardware and software support for storage, extraction, communication, computation, and load balance of nonzero and trivial elements and inputs [85]. Several techniques for exploiting structured sparsity for microcontrollers have emerged. **Bayesian compression** [86], [87] assumes hierarchical, sparsity-promoting priors on channels (output activations for convolutional layers and input features for fully connected layers) via variational inference, approximating the weight posterior by a certain distribution. For the same accuracy, Bayesian compression can reduce parameter count by 80× over unpruned models. Layerwise **SIMD-aware weight pruning** [88] divides the weights into groups equal to the SIMD width of the microcontroller for maximal SIMD unit utilization and column index sharing. Trivial weight groups are pruned based on the root mean square of each group. SIMD-aware pruning provides 3.54× speedup and 88% reduction in model size compared to 1.90× speedup and 80% reduction in model size provided by traditional weight pruning over unpruned models. **Differentiable network pruning** [89] performs structured channel pruning during training by applying channelwise binary masks depending on channel salience. The size of each layer is learned through bilevel continuous gradient descent relaxation through pruning feedback and resource feedback losses without additional training overhead. Compared to traditional pruning methods, differentiable pruning provides up to 1.7× speedup while compressing unpruned models by 80×. **Doping** [90], [91] improves the accuracy and compression factor of networks compressed using structured matrices [e.g., Kronecker products (KPs)] by adding an extremely sparse matrix using comatrix regularization to reduce comatrix adaptation during training. Doped KP matrices achieve a 2.5–5.5× speedup and 1.3–2.4× higher compression factor over traditional compression techniques, beating weight pruning and low-rank methods with 8% higher accuracy.

TABLE IV
FEATURES OF NOTABLE TINYML MODEL COMPRESSION FRAMEWORKS

| Framework | Compression Type | Parameters | Size or Latency Change* | Open-Source |
|---|---|---|---|---|
| TensorFlow Lite [46] | Post-training quantization | Bit-width (float16, int16, int8), scheme (full-integer, dynamic, float16) | 4× smaller, 2-3× speedup [47] | ✓ |
| | Quantization-aware training | Bit-width (arbitrary) | Depends on bit-width (upto 8× smaller) | |
| | Weight pruning | Sparsity distribution (constant, polynomial decay), pruning policy | 5-10× smaller [48], 4× speedup [49] | |
| | Weight clustering | Number of clusters, initial distribution (random, density-based, linear) | 3-6× smaller | |
| QKeras [50] | Quantization-aware training | Bit-width (arbitrary), symmetry, quantized layer definitions, quantized activation functions | Depends on bit-width (upto 8× smaller) | ✓ |
| Qualcomm AIMET [51] | Post-training quantization | Bit-width (arbitrary), rounding mode (nearest, stochastic), scheme (data-free, adaptive rounding) | Depends on bit-width (upto 8× smaller) | ✓ |
| | Quantization-aware training | Bit-width (arbitrary), scheme (vanilla, range-learning) | | |
| | Channel pruning | Compression ratio, layers to ignore, compression ratio candidates, reconstruction samples, cost metric | 2× smaller | |
| | Matrix factorization | Factorization algorithm (weight SVD, spatial SVD), compression ratio, fine-tuning (per layer, rank rounding) | | |
| Plumerai LARQ [52] | Binarized network training | Bit-width (int1), quantized activation functions, quantized layer definitions (convolution primitives and dense), binarized model backbones | 8× smaller, 8.5-19× speedup (with LARQ compute engine) [53] | ✓ |
| Microsoft NNI [54] | Post-training quantization | Scheme (naive, observer), bit-width (8-bit, arbitrary), type (dynamic, integer), operator type | Depends on bit-width (upto 8× smaller) | ✓ |
| | Quantization-aware training | Scheme (Vanilla, LARQ, learned step size, DoReFa), bit-width (8-bit, arbitrary), type (dynamic, integer), operator type, optimizer | | |
| | Basic pruners | Sparsity distribution, mode (normal, dependency-aware), operator type, training scheme, pruning algorithm (level, L1, L2, FPGM, slim, ADMM, activation APOZ rank, activation mean rank, Taylor FO) | 1.4-20× smaller, 1.6-5× speedup | |
| | Scheduled pruners | All parameters of basic pruners, basic pruning algorithm, scheduled pruning algorithm (linear, AGP, lottery ticket, simulated annealing, auto compress, AMC) | 1.1-120× smaller, 1.81-4× speedup | |
| CMix-NN [55] | Quantization-aware training (mixed precision) | Bit-width (int2, int4, int8), weight quantization type (per-channel, per-layer), batch normalization folding type and delay, memory constraints, quantized convolution primitives | 7× smaller | ✓ |
| Microsoft SeeDot [56] | Post-training quantization (with autotuned and optimized operators) | Bit-width (8-bit), model (Bonsai [57], ProtoNN [58], Fast-GRNN [59], RNNPool [60]), error metric, scale parameter | 2.4-82.2× speedup [56] | ✓ |
| Genesis@ [24] | Tucker decomposition and weight pruning | Rank decomposition, network configuration, sparsity distribution, pruning policy, sensing energy, communication energy | 2-109x smaller | ✗ |

*for ∼1-4% drop in accuracy over uncompressed models.
@ compression framework for intermittent computing systems.

## VI. LIGHTWEIGHT MACHINE LEARNING BLOCKS

To reduce the memory footprint and latency while retaining the performance of ML models running on microcontrollers, several ultra-lightweight ML blocks have been proposed, as illustrated in Fig. 2. We describe some of these blocks in this section.

### A. Sparse Projection

When the input feature space is high-dimensional, sparsely projecting input features onto a low-dimensional linear manifold, called prototypes, can reduce the parameter count and improve compute efficiency of models. The projection matrix can be learned as part of the model training process using stochastic gradient descent and iterative hard thresholding to mitigate accuracy loss. Bonsai [57] is a nonlinear, shallow, and sparse DT that can make inferences on prototypes. Similarly, ProtoNN [58] is a lightweight kNN classifier that operates on prototypes.

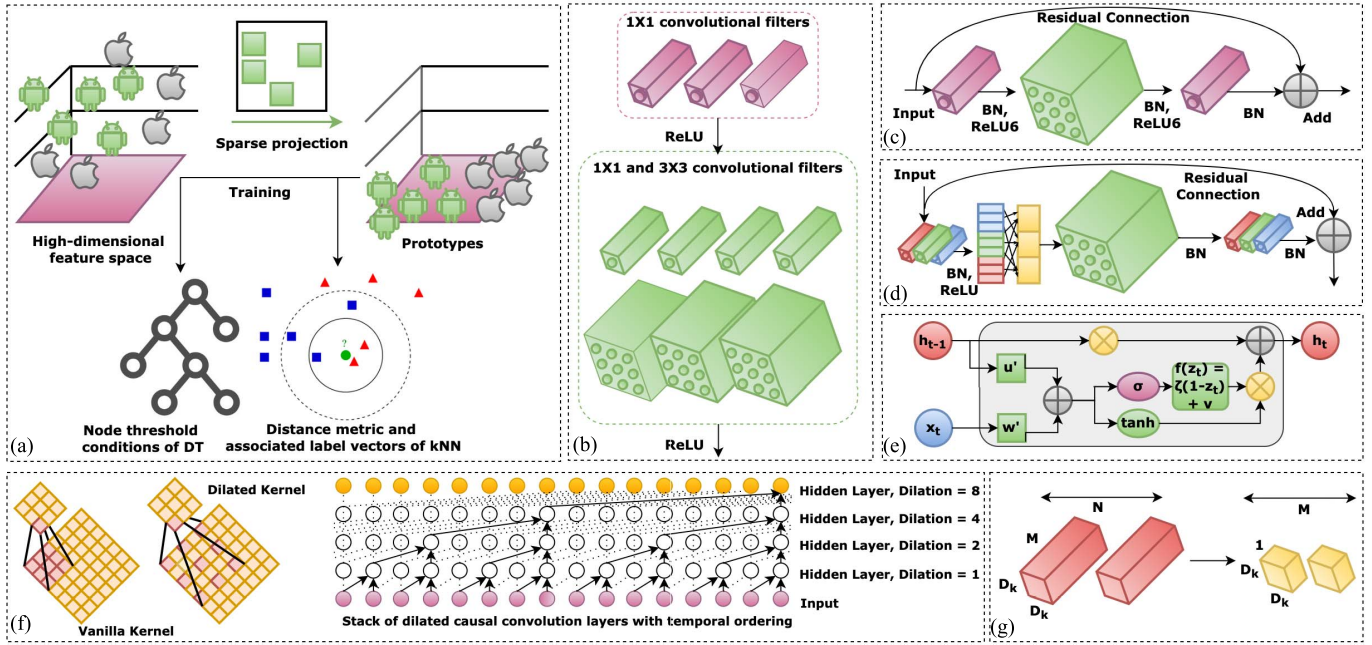

Fig. 2. Example of lightweight ML blocks. (a) Sparse projection onto low-dimensional linear manifold yields lightweight DTs and kNN classifiers. (b) Fire module containing bottleneck (pointwise) and excitation (pointwise and depthwise) convolutional layers. (c) Inverted residual connection between squeeze layers instead of excitation layers reduces memory and compute. (d) Group convolution with channel shuffle improves cross-channel relations. (e) Adding a gated residual connection and enforcing RNN matrices to be low rank, sparse, and quantized yields stable and lightweight RNN. (f) TCNs extract spatiotemporal representations using causal and dilated convolution kernels. (g) Depthwise separable convolution yields 7–9× memory savings over vanilla convolution kernel (figure adapted from [2]).

### B. Lightweight Spatial Convolution

SqueezeNet [92] brought on several microarchitectural enhancements to AlexNet [93]. These include replacing 3 × 3 kernels with pointwise filters, decreasing input channel count using pointwise filters as a linear bottleneck, and late downsampling to enhance feature maps. The resulting network consists of stacked "fire modules," with each module containing a bottleneck layer (layer with pointwise filters) and an excitation layer (mix of pointwise and 3 × 3 filters). Using pruning, quantization, and encoding, SqueezeNet reduced the size of AlexNet by 510× (<0.5 MB). MobileNetsV1 [12] introduced depthwise separable convolution [11] (channel-wise convolution followed by bottleneck layer), and width and resolution multipliers to control layer width and input resolution of AlexNet. Depthwise separable convolution is 9× cheaper and induces 7–9× memory savings over 3 × 3 kernels. MobileNetV2 [94] introduced the concepts of inverted residuals and linear bottleneck, where a residual connection exists between bottleneck layers rather than excitation layers, and a linear output is enforced at the last convolution of a residual block. To reduce channel count, the depthwise separable convolution layer can be enclosed between the pointwise group convolution layer with channel shuffle, thereby improving the semantic relation between input and output channels across all the groups through the use of wide activation maps [95]. Instead of having residual connections across two layers, the gradient highway can act as a medium to feed each layer activation maps of all preceding layers. This is known as channelwise feature concatenation [96] and encourages reuse and stronger propagation of low-complexity diversified feature maps and gradients while drastically reducing network parameter count.

### C. Lightweight Multiscale Spatial Convolution

For scalable, efficient, and real-time object detection across scales, EfficientDet [97] introduced a bidirectional feature pyramid network (FPN) to aggregate features at different resolutions with a two-way information flow. The feature network topology is optimized through NAS via heuristic compound scaling of weight, depth, and resolution. EfficientDet is 4–9× smaller, uses 13-42× fewer FLOPs, and outperforms (in terms of latency and mean average precision) YOLOv3, RetinaNet, AmoebaNet, Resnet, and DeepLabV3. Scaled-YOLOv4 [98] converts portions of FPN of YOLOv4 [99] to cross-stage partial networks [100], which saves up to 50% computational budget over vanilla CNN backbones. Removal or fusion of batch normalization layers and downscaling input resolution can speed up multiresolution inference by 3.6–8.8× [101] over vanilla YOLO [102] or MobileNetsV1 [12].

### D. Low-Rank, Stabilized, and Quantized Recurrent Models

Although RNNs are lightweight by design, they suffer from exploding and vanishing gradient problem (EVGP) for long time-series sequences. Widely used solutions to EVGP, namely, long short-term memory (LSTM) [103], gated recurrent units [104], and unitary RNN [105], either cause a loss in accuracy or increase memory and compute overhead. FastRNN [59] solves EVGP by adding a weighted residual connection with two scalars between RNN layers to stabilize gradients during training without adding significant compute overhead. The scalars control the hidden state update extent based on inputs. FastGRNN [59] then converts the residual connection to a gate while enforcing the RNN matrices to be low-rank, sparse, and quantized. The resulting RNN is $35\times$ smaller than gated or unitary RNN. Kronecker recurrent units [90], [106] use KPs to stabilize RNN training and decompose large RNN matrices into rank-preserving smaller matrices with fewer parameters, compressing RNN by 16–50× without significant accuracy loss. Doping, comatrix adaptation, and comatrix regularization can further compress Kronecker recurrent units by 1.3–2.4× [91]. Legendre memory units (LMUs) [107], derived to orthogonalize its continuous-time history, have $10\,000\times$ more capacity while being $100\times$ smaller than LSTM.

### E. Temporal Convolutional Networks

Temporal convolutional networks (TCNs) [108], [109] can jointly handle spatial and temporal features hierarchically without the explosion of parameter count, memory footprint, layer count, or overfitting. TCN convolves only on current and past elements from earlier layers but not future inputs, thereby maintaining temporal ordering without requiring recurrent connections. Dilated kernels allow the network to discover semantic connections in long temporal sequences while increasing network capacity and receptive field size with fewer parameters or layers over vanilla RNN. Two TCN layers are fused through a gated residual connection for expressive nonlinearity and temporal correlation modeling. A time-series TCN can be $100\times$ smaller over a CNN-LSTM [110], [111]. TCN also supports parallel and out-of-order training.

### F. Attention Mechanisms, Transformers, and Autoencoders

Attention mechanisms allow neural networks to focus on and extract important features from long temporal sequences. Multihead self-attention forms the central component in transformers, extracting domain-invariant long-term dependencies from sequences without recurrent units while being efficient and parallelizable [112]. Attention condensers are lightweight, self-contained, and standalone attention mechanisms independent of local context convolution kernels that learn condensed embeddings of the semantics of both local and cross-channel activations [113]. Each module contains an encoder–decoder architecture coupled with a self-attention mechanism. Coupled with machine design exploration, attention condensers have been used for image classification (4.17× fewer parameters than MobileNetsV1) [114], keyword spotting (up to 507× fewer parameters over previous work) [113], and semantic segmentation (72× fewer parameters over RefineNet and EdgeSegNet) [115] at the edge. Long-short range attention (LSRA) uses two heads (convolution and attention) to capture both local and global contexts, expanding the bottleneck while using condensed embeddings to reduce computation cost [116]. Combined with pruning and quantization, LSRA transformers can be 18× smaller than the vanilla transformer architecture. MobileViT combines the benefits of convolutional networks and transformers by replacing local processing in convolution with global processing, allowing lightweight and low-latency transformers to be implemented using convolution [117]. Instead of using special attention and transformer blocks, transformer knowledge distillation teaches a small transformer to mimic the behavior of a larger transformer, allowing up to 7.5× smaller and 9.4× faster inference over bidirectional encoder representations from transformers [118]. Customized data layout and loop reordering of each attention kernel, coupled with quantization, have allowed porting transformers onto microcontrollers [119] by minimizing computationally intensive data marshaling operations. The use of depthwise and pointwise convolution has been shown to yield autoencoder architectures as small as 2.7 kB for anomaly detection [120].

## VII. NEURAL ARCHITECTURE SEARCH

NAS is the automated process of finding the most optimal neural network within a neural network search space given target architecture and network architecture constraints, achieving a balance between accuracy, latency, and energy usage [125], [126], [127]. Table V compares several NAS frameworks developed for microcontrollers. There are three key elements in a hardware-aware NAS pipeline, namely, the search space formulation (see Section VII-A), search strategy (see Section VII-B), and cost function (see Section VII-C).

### A. Search Space Formulation

The search space provides a set of ML operators, valid connection rules, and possible parameter values for the search algorithm to explore. The neural network search space can be represented as layerwise, cellwise, or hierarchical [125].

*1) Layerwise:* In layerwise search spaces, the entire model is generated from a collection of serialized or sequential neural operators. The macroarchitecture (e.g., number of layers and dimensions of each layer), initial, and terminal layers of the network are fixed, while the remaining layers are optimized. The structure and connectivity among various operators are specified using variable-length strings, encoded in the action space of the search strategy [126]. Although such search spaces are expressive, layerwise search spaces are computationally expensive, require hardcoding associations among different operators and parameters, and are not gradient friendly.

*2) Cellwise:* For cellwise (or templatewise) search spaces, the network is constructed by stacking repeating fixed blocks or motifs called cells. A cell is a directed acyclic graph constructed from a collection of neural operators, representing some feature transformation. The search strategy finds the most optimal set of operators to construct the cell recursively in stages by treating the output of past hidden layers as hidden states to apply a predefined set of ML operations on [128]. Cell-based search spaces are more time-efficient compared to layerwise approaches and easily transferable across datasets but are less flexible for hardware specialization. In addition, the global architecture of the network is fixed.

TABLE V
NAS FRAMEWORKS TARGETTED TOWARD MICROCONTROLLERS

| Framework | Search Strategy | Hardware Profiling | Inference Engine | Optimization Parameters | Open-Source |
|---|---|---|---|---|---|
| SpArSe [86] | Gradient-driven Bayesian | Analytical | uTensor | Error, SRAM, Flash | ✗ |
| MCUNet [31] [121] | Evolutionary | Lookup tables, prediction models | TinyEngine (closed-source) | Latency, Error, SRAM, Flash | ✗ |
| MicroNets [8] | Gradient-driven | Analytical | Tensorflow Lite Micro | Latency, Error, SRAM, Flash | ✗ |
| $\mu$NAS [122] | Evolutionary | Analytical | Tensorflow Lite Micro | Latency, Error, SRAM, Flash | ✓ |
| THIN-Bayes [123] | Gradient-free Bayesian | Hardware-in-the-loop, analytical | Tensorflow Lite Micro | Latency, Error, SRAM, Flash, Arena size, Energy | ✓ |
| iNAS [124] | Reinforcement Learning | Lookup tables, analytical | Accelerated Intermittent Inference (custom) | Latency*, Error, Volatile Buffer, Flash, Power-Cycle Energy@ | ✓ |

\* sum of progres preservation, progress recovery, battery recharge and compute cost
@ sum of progres preservation, progress recovery, and compute cost

*3) Hierarchical:* In tree-based search spaces, bigger blocks encompassing specific cells are created and optimized after cellwise optimization. Primitive templates that are known to perform well are used to construct larger network graphs and higher level motifs recursively, with feature maps of low-level motifs being fed into high-level motifs. Factorized hierarchical search spaces allow each layer to have different blocks without increasing the search cost while allowing for hardware specialization [129].

For applications with extreme memory and energy budget (e.g., intermittent computing systems), the search space goes down to the execution level to include operator and inference optimizations (e.g., loop transformations, data reuse, and choice of in-place operators) rather than optimizing the model at the architectural level. iNAS [124] uses reinforcement learning (RL) to optimize the tile dimensions per layer, loop order in each layer, and the number of tile outputs to preserve in a power cycle for convolutional models. When combined with appropriate power-cycle energy, memory, and latency constraints, iNAS reduced intermittent inference latency by 60% compared to NAS frameworks operating at the architectural level with a 7% increase in search overhead. Likewise, micro-TVM [130] uses a learning-enabled schedule explorer to perform automated operator and inference optimizations at the execution level. We discuss some of these optimizations in Section VIII-A and operation of micro-TVM in Section VIII-B.

### B. Search Strategy

The search strategy involves sampling, training, and evaluating candidate models from the search space with the goal of finding the best performing model. This is done using RL, one-shot gradient-driven NAS, evolutionary algorithms (with weight sharing), or Bayesian optimization [134]. Recent techniques, known as training-free NAS, aim to perform NAS without the costly inner loop training [135].

TABLE VI
COMPARISON OF DIFFERENT NAS SEARCH STRATEGIES [5], [131]

| Search Strategy | Top-1% Accuracy | Latency$^\wedge$ | Model Size (MAC) | Training Cost (GPU hours) | Search Cost (GPU hours) |
|---|---|---|---|---|---|
| Reinforcement Learning$^\vee$ | 74%-75.2% | 58mS-70 mS | 219M-564M | None*, 180N@ | 40000N-48000N*, None@ |
| Gradient-driven | 73.1%-74.9% | 71mS | 320M-595M | 250N-384N | 96N-(288+24N) |
| Evolutionary | 72.4%-80.0% | 58mS-59mS | 230M-595M | 1200-(1200+kN) | 40 |
| Bayesian$^\vee$ | 73.4%-75.8% | - | 225M | None | 23N-552N |

dataset: ImageNet-1000, backbone network: MBNetv3, k = fine-tuning epoch count
$^\wedge$ on Google Pixel1 smartphone, N = Number of deployment scenarios for which different models must be found [5]
$^\vee$ Techniques based on RL and Bayesian usually have coupled training and search (training cost included with search cost)
\* NASNet [128] and MNASNet [132], @ MBNetV3 Search [133]

Table VI compares the performance of different NAS search strategies on the ImageNet dataset for MBNetv3 [133] backbone. We distill the insights from Table VI.

*1) Reinforcement Learning:* RL techniques, such as NASNet [128] and MNASNet [132], model NAS as a Markov Decision Process on a proxy dataset to reduce search time. RL controllers (e.g., RNNs trained via proximal policy optimization (PPO), deep deterministic policy gradient (DDPG), and Q-learning) are used to find the optimal combination of neural network cells from a predefined set recursively. The network graph can either consist of a series of repeatable and identical blocks (e.g., convolutional cells) whose structures are found via the controller or represented in a factorized hierarchical fashion via a layerwise stochastic supernetwork. Device constraints are included in the reward function to formulate a multiobjective optimization problem. Among the different RL controllers, Q-learning-based algorithms work for simple search space (i.e., discrete and finite with tens of parameters) [136] created through expert knowledge. PPO and DDPG are useful when the search space is complex (i.e., continuous with thousands of parameters) [137]. PPO-based on-policy algorithms are more stable than DDPG but demand more samples to converge than DDPG [138]. Overall, RL processes are slow to converge, preventing fast exploration of the search space. In addition, fine-tuning candidate networks increases search costs.

*2) Gradient-Driven NAS:* Differentiable NAS using continuous gradient descent relaxation can reduce the search and training cost further on the target dataset over RL-based techniques. The goal is to learn the weights and architectural encodings through a nested bilevel optimization problem with the gradients obtained approximately. The optimization problem can be efficiently handled using path binarization, where the weights and encodings of an overparametrized network are alternatively frozen during gradient update using binarized gates. The final subnetwork is obtained using path-level pruning. Hardware metrics are converted to a gradient-friendly continuous model before being used in the optimization function. The search space can consist of static blocks of directed acyclic graphs containing network weights, edge operations, activations, and hyperparameters or a factorized hierarchical supernetwork. Examples of gradient-driven NAS include DARTS [139], FBNet [140], ProxylessNAS [129], and MicroNets [8]. Drawbacks include high GPU memory consumption and training time due to large supernetwork parameter count and inability to generalize across a broad spectrum of the target hardware, requiring the NAS process to be repeated for new hardware.

*3) Evolutionary Search With Weight Sharing:* To eliminate the need for performing NAS for different hardware platforms separately and reduce the training time of candidate networks, several weight-sharing mechanisms (WS-NAS) have emerged [5], [31], [121], [143], [144]. WS-NAS decouples training from search by training a "once-for-all" supernetwork consisting of several subnetworks, which fits the constraints of eclectic target platforms. Evolutionary search is used during the search phase, where the best performing subnetworks are selected from the supernetwork, crossed, and mutated to produce the next generation of candidate subnetworks. Progressive shrinking and knowledge distillation ensure all the subnetworks are jointly fine-tuned without interfering with large subnetworks. Evolutionary search can also be applied to RL search spaces [145] for faster convergence or applied on several candidate architectures not part of a supernetwork [122]. Nevertheless, evolutionary WS-NAS suffers from excessive computation and time requirements due to supernetwork training, exacerbated by fine-tuning of candidate networks and slow convergence of evolutionary algorithms.

*4) Bayesian Optimization:* When training infrastructure is weak, the search space and hardware metrics are discontinuous, and the training cost per model is high, Bayesian NAS is used as a black-box optimizer. Given their problem-independent nature, Bayesian NAS can be applied across different datasets and heterogenous architectures without being constrained to one specific type of network (e.g., CNN or RNN), provided support for conditional search. The performance of the optimizer is highly dependent on the surrogate model [146]. The most widely adopted surrogate model is the Gaussian process, which allows uncertainty metrics to propagate forward while looking for a Pareto-optimal frontier and is known to outperform other choices, such as random forest or tree of Parzen estimators [146]. The acquisition function decides the next set of parameters from the search space to sample from, balancing exploration and exploitation. The loss function is modeled as a constrained single-objective or scalarized multiobjective optimization problem. Examples include SpArSe [86], Vizier [147], and THIN-Bayes [123]. Unfortunately, Bayesian NAS does not perform well in high-dimensional search spaces (e.g., performance degrades beyond a dozen parameters [148]). Moreover, Bayesian optimizers are typically used to optimize hyperparameters for fixed network architectures instead of multiple architectures as the Gaussian process does not directly support conditional search across architectures. Only THIN-Bayes can sample across different architectures thanks to support for conditional search via multiple Gaussian surrogates [123].

*5) Training-Free NAS:* Training-free NAS estimates the accuracy of a neural network either by using proxies developed from architectural heuristics of well-known network architectures [135] or by using a graph neural network (GNN) to predict the accuracy of models generated from a known search space [149], [150]. Examples of gradient-based accuracy proxies include the correlation of ReLU activations (Jacobian covariance) between minibatch datapoints at CNN initialization [151], the sum of the gradient Euclidean norms after training with a single minibatch datapoints [152], change in loss due to layer-level pruning after training with a single minibatch datapoint (Fisher) [152], change in loss due to parameter pruning after training with a single minibatch datapoint (Snip) [153], change in gradient norm due to parameter pruning after training with a single minibatch datapoint (Grasp) [154], the product of all network parameters (synaptic flow) [155], the spectrum of the neural tangent kernel [156], and the number of linear regions in the search space [156]. Gradient-based proxies still require the use of a GPU for gradient calculation. Recently, Li et al. [135] proposed a gradient-free accuracy proxy, namely, the sum of the average degree of each building block in a CNN from a network topology perspective. Unfortunately, both proxies and GNN suffer from the lack of generalizability across different datasets, model architectures, and design spaces, while the latter also suffers from the training cost of the accuracy prediction network itself.

### C. Cost Function

The cost function provides numerical feedback to the search strategy about the performance of a candidate network. Common parameters in the cost function include network accuracy, SRAM usage, flash usage, latency, and energy usage. The goal of NAS is to find a candidate network that finds the extrema of the cost function, i.e., the cost function can be thought of as seeking a Pareto-optimal configuration of network parameters.

### D. Cost Function Formulation

The cost function can be formulated as either a single or multiobjective optimization problem. Single-objective optimization problems only optimize for model accuracy. To take hardware constraints into account, single-objective optimization problems are usually treated as constrained optimization problems with hardware costs acting as regularizers [123]. Multiobjective cost functions are usually transformed into a single-objective optimization problem via weighted-sum or scalarization techniques [86] or solved using genetic algorithms.

### E. Hardware Profiling

Hardware-aware NAS employs hardware-specific cost functions or search heuristics via hardware profiling. The target hardware can be profiled in real time by running sampled models on the actual target device (hardware-in-the-loop), estimated using lookup tables, prediction models, and silicon-accurate emulators [157] or analytically estimated using architectural heuristics. Common hardware profiling techniques are shown in Table VII. Hardware-in-the-loop is slowest but most accurate during NAS runtime, while analytical estimation is fastest but least accurate [125], [134]. Examples of analytical models for microcontrollers include using FLOPs as a proxy for latency [8], [122] and standard RAM usage model [86] for working memory estimation. Recently, latency prediction models have been made more accurate through kernel (execution unit) detection and adaptive sampling [158]. For intermittent computing systems, the latency is the time required for progress preservation (writing progress indicators and computed tile outputs to flash at the end of a power cycle), progress recovery (system reboot, loading progress indicators, and tiled outputs into SRAM), battery recharge, and running inference (cost of computing multiple tiles per energy cycle) [124]. The SRAM usage in such systems is the sum of memory consumed by the input feature map, weights, and output feature map, dependent upon the tile dimensions, loop order, and preservation batch size in the search space [124].

TABLE VII
NAS HARDWARE PROFILING STRATEGIES FOR MICROCONTROLLERS

| Method | Speed | Accuracy | NAS Frameworks |
|---|---|---|---|
| Real measurements | Slow | High | THIN-Bayes [123], MNASNet [132], One-shot NAS [141] |
| Lookup tables | Fast-Medium | Medium-High | FBNet [140], Once-for-All [5], MCUNet [31] [121] |
| Prediction models | Medium | Medium | ProxylessNAS [129], Once-for-All [5], MCUNet [31] [121], LEMONADE [142] |
| Analytical | Fast | Low | THIN-Bayes [123], MicroNets [8], $\mu$NAS [122], SpArSe [86] |

## VIII. TINYML SOFTWARE SUITES

After the best model is constructed from lightweight ML blocks through NAS, the model needs to be prepared for deployment onto microcontrollers. This is performed by TinyML software suites that generate embedded code and perform operator and inference engine optimizations, some of which are shown in Fig. 3 and discussed in Section VIII-A. In addition, some of these frameworks also provide inference engines for resource management and model execution during deployment. We discuss features of notable TinyML software suites in Section VIII-B.

### A. Operator and Inference Optimizations

All TinyML software suites perform several operator and inference engine optimizations to improve data locality, memory usage, and spatiotemporal execution [162]. Common techniques include the use of fused or in-place operators [130], loop transformations [161], and data reuse (output sharing or value sharing) [163].

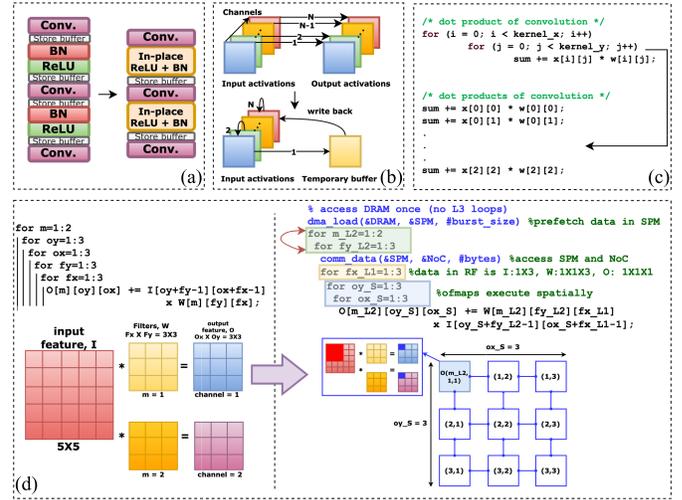

Fig. 3. Example operator optimizations performed by TinyML software suites. (a) Use of fused and in-place activated operators reduces memory access cost and improves inference speed [158], [159]. (b) Converting depthwise convolution to in-place depthwise convolution reduces peak memory usage by 1.6× by allowing the first channel output activation (stored in a buffer) to overwrite the previous channel's input activation until written back to the last channel's input activation [31]. (c) Loop unrolling eliminates branch instruction overhead [31]. (d) Loop tiling encourages reuse of array elements within each tile by partitioning the loop's iterative space into blocks [160], while loop reordering (with tiling) improves spatiotemporal execution and locality of reference within device memory constraints [161], [162].

*1) In-Place and Fused Operators:* Operator fusion or folding combines several ML operators into a specialized kernel without saving the intermediate feature representations in memory (known as in-place activation) [130]. The software suites follow user-defined rules for operator fusion depending on graph operator type (e.g., injective, reduction, complex-out fusable, and opaque) [130]. The use of fused and in-place operators has been shown to reduce memory usage by 1.6× [31] and improve speedup by 1.2–2× [130].

*2) Loop Transformations:* Loop transformations aim to improve spatiotemporal execution and inference speed by reducing loop overheads [161]. Common loop transformations include loop reordering, loop reversal, loop fusion, loop distribution, loop unrolling, and loop tiling [160], [161], [162]. Loop reordering (and reversal) finds the loop permutation that maximizes data reuse and spatiotemporal locality. Loop fusion combines different loop nests into one, thereby improving temporal locality, and increasing data locality and reuse by creating perfect loop nests from imperfect loop nests. To enable loop permutation for loop nests that are not permutable, loop distribution breaks a single loop into multiple loops [161]. Loop unrolling helps eliminate branch penalties and helps hide memory access latencies [130]. Loop tiling improves data reuse by diving the loops into blocks while considering the size of each level of memory hierarchy [160].

*3) Data Reuse:* Data reuse aims to improve data locality and reduce memory access costs. While data reuse is mostly achieved through loop transformations, several other techniques have also been proposed. CMSIS-NN provides special pooling and multiplication operations to promote

TABLE VIII
FEATURES OF NOTABLE TinyML SOFTWARE SUITES FOR MICROCONTROLLERS

| Framework | Supported Platforms | Supported Models | Supported Training Libraries | Open-Source | Free |
|---|---|---|---|---|---|
| TensorFlow Lite Micro (Google) [167] [46] | ARM Cortex-M, Espressif ESP32, Himax WE-I Plus | NN | TensorFlow | ✓ | ✓ |
| uTensor (ARM) [168] | ARM Cortex-M (Mbed-enabled) | NN | TensorFlow | ✓ | ✓ |
| uTVM (Apache) [130] | ARM Cortex-M | NN | PyTorch, TensorFlow, Keras | ✓ | ✓ |
| EdgeML (Microsoft) [57] [58] [59] [60], [169]–[171] | ARM Cortex-M, AVR RISC | NN, DT, kNN, unary classifier | PyTorch, TensorFlow | ✓ | ✓ |
| CMSIS-NN (ARM) [164] | ARM Cortex-M | NN | PyTorch, TensorFlow, Caffe | ✓ | ✓ |
| EON Compiler (Edge Impulse) [32] | ARM Cortex-M, TI CC1352P, ARM Cortex-A, Espressif ESP32, Himax WE-I Plus, TENSAI SoC | NN, k-means, regressors (supports feature extraction) | TensorFlow, Scikit-Learn | ✗ | ✓ |
| STM32Cube.AI (STMicroelectronics) [172] | ARM Cortex-M (STM32 series) | NN, k-means, SVM, RF, kNN, DT, NB, regressors | PyTorch, Scikit-Learn, TensorFlow, Keras, Caffe, MATLAB, Microsoft Cognitive Toolkit, Lasagne, ConvnetJS | ✗ | ✓ |
| NanoEdge AI Studio (STMicroelectronics) [173] | ARM Cortex-M (STM32 series) | Unsupervised learning | - | ✗ | ✗ |
| EloquentML [72] | ARM Cortex-M, Espressif ESP32, Espressif ESP8266, AVR RISC | NN, DT, SVM, RF, XGBoost, NB, RVM, SEFR (feature extraction through PCA) | TensorFlow, Scikit-Learn | ✓ | ✓ |
| Sklearn Porter [174] | - | NN (MLP), DT, SVM, RF, AdaBoost, NB | Scikit-Learn | ✓ | ✓ |
| EmbML [175] | ARM Cortex-M, AVR RISC | NN (MLP), DT, SVM, regressors | Scikit-Learn, Weka | ✓ | ✓ |
| FANN-on-MCU [176] | ARM Cortex-M, PULP | NN | FANN | ✓ | ✓ |
| SONIC, TAILS[@] [24] | TI MSP430 | NN | TensorFlow | ✓ | ✓ |

[@] inference framework for intermittent computing systems.

data reuse [164]. TF-Net [163] proposed the use of direct buffer convolution on Cortex-M microcontrollers to reduce input unpacking overhead, which reuses inputs in the current window unpacked in a buffer space for all weight filters. Input reuse reduces SRAM usage by 2.57× and provides 2× speedup. Similarly, for GAP8 processors, the PULP-NN library provides a reusable im2col buffer (height–width–channel data layout) to reduce im2col creation overhead [165], [166], providing partial spatial data reuse. PULP-NN also features register-level data reuse, achieving 20% speedup over CMSIS-NN and 1.9× improvement over native GAP8-NN libraries.

### B. Notable TinyML Software Suites

Notable open-source TinyML frameworks and inference engines include TensorFlow Lite Micro (TFLM) [46], [167], uTensor [168], uTVM [130], Microsoft EdgeML [57], [58], [59], [60], [169], [170], [171], CMSIS-NN [164], EloquentML [72], Sklearn Porter [174], EmbML [175], and FANN-on-MCU [176]. Closed-source TinyML frameworks and inference engines include STM32Cube.AI [172], NanoEdge AI Studio [173], Edge Impulse EON Compiler [32], TinyEngine [31] [121], Qeexo AutoML [35], Deeplite Neutrino [177], Imagimob AI [178], Neuton TinyML [179], Reality AI [180], and SensiML Analytics Studio and Knowledge Pack [34]. Table VIII compares the features of some of these frameworks.

*1) TensorFlow Lite Micro:* TFLM [46], [167] is a specialized version of TFLite aimed toward optimizing TF models for Cortex-M and ESP32 MCU. TFLite Micro embraces several embedded runtime design philosophies. TFLM drops uncommon features, data types, and operations for portability.

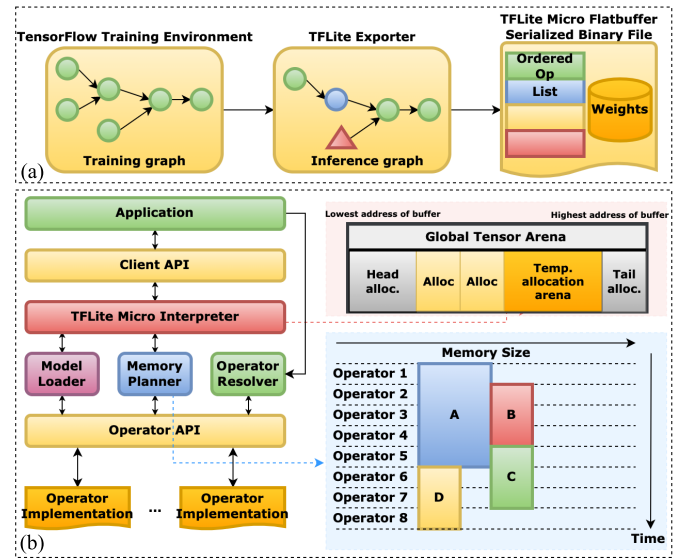

Fig. 4. Operation of TFLM, an interpreter-based inference engine. (a) Training graph is frozen, optimized, and converted to a flatbuffer serialized model schema, suitable for deployment in embedded devices. (b) TFLM runtime API preallocates a portion of memory in the SRAM (called arena) and performs bin-packing during runtime to optimize memory usage (figure adapted from [167]).

It also avoids specialized libraries, operating systems, or build-system dependencies for heterogeneous hardware support and memory efficiency. TFLM avoids dynamic memory allocation to mitigate memory fragmentation. TFLM interprets the neural network graph at runtime rather than generating C++ code to support easy pathways for upgradability, multitenancy, multithreading, and model replacement while sacrificing finite savings in memory. Fig. 4 summarizes the operation of TFLM.

TFLM consists of three primary components. First, the **operator resolver** links only essential operations to the model binary file. Second, TFLM preallocates a contiguous memory stack called the **arena** for initialization and storing runtime variables. TFLM uses a two-stack allocation strategy to discard initialization variables after their lifetime, thereby minimizing memory consumption. The space between the two stacks is used for temporary allocations during memory planning, where TFLM uses bin-packing to encourage memory reuse and yield optimal compacted memory layouts during runtime. Finally, TFLM uses an **interpreter** to resolve the network graph at runtime, allocate the arena, and perform runtime calculations. TFLM was shown to provide $2.2\times$ speedup and $1.08\times$ memory and flash savings over CMSIS-NN for image recognition [31].

*2) uTensor:* uTensor [168] generates C++ files from TF models for Mbed-enabled boards, aiming to generate models of <2 kB in size. It is subdivided into two parts. The **uTensor core** provides a set of optimized runtime data structures and interfaces under computing constraints. The **uTensor library** provides default error handlers, allocators, contexts, ML operations, and tensors built on the core. Basic data types include integral type, uTensor strings, tensor shape, and quantization primitives borrowed from TFLite. Interfaces include the memory allocator interface, tensor interface, tensor maps, and operator interface. For memory allocation, uTensor uses the concept of arena borrowed from TFLM. In addition, uTensor boasts a series of optimized (built to run CMSIS-NN under the hood), legacy, and quantized ML operators consisting of activation functions, convolution operators, fully connected layers, and pooling.

*3) uTVM:* Micro-TVM [130] extends the TVM compiler stack to run models on bare-metal IoT devices without the need for operating systems, virtual memory, or advanced programming languages. Micro-TVM first generates a high-level and quantized computational graph (with support for complex data structures) from the model using the **relay module**. The functional representation is then fed into the TVM **intermediate representation module**, which generates C-code by performing operator and loop optimizations via AutoTVM and Metascheduler, procedural optimizations, and graph-level modeling for whole program memory planning. AutoTVM consists of an automatic **schedule explorer** to generate promising and valid operator and inference optimization configurations for a specific microcontroller and an XGBoost model to predict the performance of each configuration based on features of the lowered loop program. The developer can either specify the configuration parameters to explore using a schedule template specification API, or possible parameters can be extracted from the hardware computation description written in the tensor expression language. AutoTVM has lower data and exploration costs than black-box optimizers (e.g., ATLAS [181]) and provides more accurate modeling than polyhedral methods [182] without needing a hardware-dependent cost model. The generated code is integrated alongside the TVM C runtime, built, and flashed onto the device. The inference is made on the device using a graph extractor. AutoTVM was shown to generate code that is only $1.2\times$ slower compared to handcrafted CMSIS-NN-based code for image recognition.

*4) Microsoft EdgeML:* EdgeML provides a collection of lightweight ML algorithms, operators, and tools aimed toward deployment on Class 0 devices, written in PyTorch and TF. Included algorithms include Bonsai [57], ProtoNN [58], FastRNN [59], FastGRNN [59], ShallowRNN [169], EMI-RNN [170], RNNPool [60], and DROCC [171]. EMI-RNN exploits the fact that only a small, tight portion of a time-series plot for a certain class contributes to the final classification, while other portions are common among all classes. Shallow-RNN is a hierarchical RNN architecture that divides the time-series signal into various blocks and feeds them in parallel to several RNNs with shared weights and activation maps. RNNPool is a nonlinear pooling operator that can perform "pooling" on intermediate layers of a CNN by a downsampling factor much larger than 2 ($4-8\times$) without losing accuracy while reducing memory usage and decreasing compute. A deep robust one-class classifier (DROCC) is an OCC under limited negatives and anomaly detector without requiring domain heuristics or handcrafted features. The framework also includes a quantization tool called SeeDot [56].

*5) CMSIS-NN:* Cortex Microcontroller Software Interface Standard-NN [164] was designed to transform TF, PyTorch, and Caffe models for Cortex-M series MCU. It generates C++ files from the model, which can be included in the main program file and compiled. It consists of a collection of optimized neural network functions with fixed-point quantization, including fully connected layers, depthwise separable convolution, partial image-to-column convolution, in situ split x-y pooling, and activation functions (ReLU, sigmoid, and tanh, with the latter two implemented via lookup tables). It also features a collection of support functions including data type conversion and activation function tables (for sigmoid and tanh). CMSIS-NN provides $4.6\times$ speedup and $4.9\times$ energy savings over nonoptimized convolutional models.

*6) Edge Impulse EON Compiler:* Edge Impulse [32] provides a complete end-to-end model deployment solution for TinyML devices, starting with data collection using IoT devices, extracting features, training the models, and then deployment and optimization of models for TinyML devices. It uses the interpreter-less **edge optimized neural (EON) compiler** for model deployment while also supporting TFLM. EON compiler directly compiles the network to C++ source code, eliminating the need to store ML operators that are not in use (at the cost of portability). An EON compiler was shown to run the same network with 25%–55% less SRAM and 35% less flash than TFLM.

*7) STM32Cube.AI and NanoEdge AI Studio:* X-Cube-AI from STMicroelectronics [172] generates STM32 compatible C code from a wide variety of deep-learning frameworks (e.g., PyTorch, TensorFlow, Keras, Caffe, MATLAB, Microsoft Cognitive Toolkit, Lasagne, and ConvnetJS). It allows quantization (min–max), operator fusion, and the use of external flash or SRAM to store activation maps or weights. The tool also features functions to measure system performance and deployment accuracy and suggests a list of compatible STM32 platforms based on model complexity.

TABLE IX
FEATURES OF NOTABLE TINYML ON-DEVICE LEARNING FRAMEWORKS

| Framework | Working Principle | Supported Hardware | Tested Application | Network Type | Open-source |
|---|---|---|---|---|---|
| Learning in the Wild [183] | W: Per-output feature distribution divergence. H: Transfer learning on last-layer; sample importance weighing to maximize learning effect. T: Gradient norm for sample selection via uncertainty and diversity. | TI MSP430 | Image recognition (MNIST, CIFAR-10, GT-SRB) | CNN | ✗ |
| TinyOL [184] | W: Running mean and variance of streaming input H: Transfer learning on additional layer at the output of the frozen network using stochastic gradient descent (SGD). | ARM Cortex-M | Anomaly detection | Autoencoder | ✗ |
| ML-MCU [185] | H: Optimized SGD (inherits stability of GD and efficiency of SGD); optimized one-versus-one (OVO) binary classifiers for multiclass classification | ARM Cortex-M, Espressif ESP32 | Image recognition (MNIST, mHealth (Heart Disease, Breast Cancer), Other (Iris) | Optimized OVO binary classifiers | ✓ |
| Train++ [186] | W: Confidence score of prediction. H: Incremental training via constrained optimization classifier update | ARM Cortex-M, ARM Cortex-A, Espressif ESP32, Xtensa LX | Image recognition (MNIST, Banknote Authentication), mHealth (Heart Disease, Breast Cancer, Haberman's Survival), Other (Iris, Titanic Survival) | Binary classifiers | ✓ |
| TinyTL [19] | H: Update bias instead of weights and use lite residual learning modules to recoup accuracy loss | ARM Cortex-A | Face recognition (CelebA), Image recognition (Cars, Flowers, Aircraft, CUB, Pets, Food, CIFAR-10, CIFAR-100) | CNN (ProxylessNAS-MB, MBNetV2) | ✓ |
| Imbal-OL [187] | T: Weighted replay and oversampling for minority classes | ARM Cortex-A | Image recognition (CIFAR-10, CIFAR-100) | CNN (ResNet-18) | ✗ |
| QLR-CL [188] | H: Continual learning with quantized latent replays (store activation maps at latent replay layer instead of samples), slow-learning below the latent replay layer. | PULP | Image recognition (Core50) | CNN (MBNetV1) | ✗ |

W: When to learn, H: How to learn, T: What to learn (sample selection)

X-Cube-AI was shown to provide $1.3\times$ memory reduction and $2.2\times$ speedup over TFLM for gesture recognition and keyword spotting [189]. NanoEdge AI Studio [173] is another AutoML framework from STMicroelectronics for prototyping anomaly detection, outlier detection, classification, and regression problems for STM32 platforms, including an embedded emulator.

*8) Eloquent MicroML and TinyML:* MicroMLgen ports DTs, support vector machines (linear, polynomial, radial kernels, or one-class), random forests, XGboost, Naive Bayes, relevant vector machines, and SEFR (a variant of SVM) from SciKit-Learn to Arduino-style C code, with the model entities stored on flash. It also supports onboard feature projection through PCA. TinyMLgen ports TFLite models to optimized C code using TFLite's code generator [72].

*9) Sklearn Porter:* Sklearn Porter [174] generates C, Java, PHP, Ruby, GO, and Javascript code from Scikit-Learn models. It supports the conversion of support vector machines, DTs, random forests, AdaBoost, kNNs, Naive Bayes, and multilayer perceptrons.

*10) EmbML:* Embedded ML [175] converts logistic regressors, DTs, multilayer perceptrons, and support vector machine (linear, polynomial, or radial kernels) models generated by Weka or Scikit-Learn to C++ code native to embedded hardware. It generates initialization variables, structures, and functions for classification, storing the classifier data on flash to avoid high memory usage, and supports the quantization of floating-point entities. EmbML was shown to reduce memory usage by 31% and latency by 92% over Sklearn Porter.

*11) FANN-on-MCU:* FANN-on-MCU [176] ports multilayer perceptrons generated by fast artificial neural network (FANN) library to Cortex-M series processors. It allows model quantization and produces an independent callable C function based on the specific instruction set of the MCU. It takes the memory of the target architecture into account and stores network parameters in either RAM or flash depending upon whichever does not overflow and is closer to the processor (e.g., RAM is closer than flash).

*12) SONIC and TAILS:* Software-only neural intermittent computing (SONIC) and tile-accelerated intermittent low-energy accelerator (LEA) support (TAILS) [24] are inference engines for intermittent computing systems. SONIC eliminates redo-logging, task transitions, and wasted work associated with moving data between SRAM and flash by introducing **loop continuation**, which allows loop index modification directly on the flash without expensive saving and restoring. To ensure idempotence, SONIC uses **loop ordered buffering** (loop reordering and double buffering partial feature maps to eliminate commits in a loop iterations) and **sparse undo-logging** (buffer reuse to ensure idempotence for sparse ML operators). SONIC introduces a latency overhead of only 25%–75% over nonintermittent neural network execution (compared to $10\times$ overhead from baseline intermittent model execution frameworks), reducing inference energy by $6.9\times$ over competing baselines. TAILS exploits LEA in MSP430 microcontrollers to maximize throughput using direct-memory access and parallelism. LEA supports the acceleration of finite-impulse-response discrete-time convolution. TAILS further reduces inference energy by $12.2\times$ over competing baselines.

## IX. ONLINE LEARNING

After deployment, the on-device model needs to be periodically updated to account for shifts in feature distribution in the wild [183]. While models trained on new data on a server could be sent out to the microcontroller once in a while, limited communication bandwidth and privacy concerns can prevent offloading the training to a server. However, the conventional training memory and energy footprint are much larger than the inference memory and energy footprint, rendering traditional GPU-based training techniques unsuitable for microcontrollers [19]. Thus, several on-device training and FL frameworks have emerged for microcontrollers, as summarized in Tables IX and X.

TABLE X
FEATURES OF NOTABLE TINYML FL FRAMEWORKS

| Framework | FL Strategy | Communication Stack | Scalability and Heterogeneity | Privacy | Client Hardware (language) | Open-source |
|---|---|---|---|---|---|---|
| Flower [190] [191] | FedAvg, Fault tolerant FedAvg, FedProx, QFedAvg, FedAdagrad, FedYogi, FedAdam | Bidirectional gRPC and *ClientProxy* (language, communication and serialization agnostic) | FedFS (partial work, importance sampling, and dynamic timeouts to handle bandwidth heterogeneity); Virtual Client Engine for scheduling and resource management (15M clients tested) | Salvia secure aggregation | CPU, GPU, MCU (Python, Java, C++) | ✓ |
| FedPARL [192] | Reparametrized FedAvg with sample-based pruning | None (simulated) | Resource tracking (memory, battery life, bandwidth, and data volume); Trust value tracking (task completion, delay, model integrity); Partial work (12 clients tested) | Vanilla model aggregation | None (simulated) | ✗ |
| DIoT [193] | FedAvg | Bidirectional WebSocket protocol over WiFi and Ethernet | AuDI device-type identification (15 clients tested) | Vanilla model aggregation | CPU, GPU (Python and JavaScript) | ✗ |
| PruneFL [194] | FedAvg with adaptive and distributed pruning | WiFi and Ethernet, with distributed pruning to reduce communication overhead | Adaptive pruning to modify local models based on resource availability (9 clients tested) | Vanilla model aggregation | CPU, MCU (Python) | ✓ |
| TinyFedTL [195] | FedAvg with last layer transfer learning | USART | 9 clients tested | Vanilla model aggregation | MCU (C++) | ✓ |
| FLAgr [196] | Reinforcement learning | None (simulated) | Real-time collaboration scheme discovery via deep deterministic policy gradient (1000 clients tested) | Rating feedback mechanism | None (simulated) | ✗ |
| PerFit [197] | FedPer, FedHealth, FedAvg, Personalized FedAvg, MOCHA, FedMD, Federated Distillation | WiFi, BLE, Cellular (simulated) | Federated transfer learning, federated distillation, federated meta-learning, and federated multi-task learning to personalize the model, device and statistical heterogeneity (30 clients tested) | Vanilla model aggregation | None (simulated) | ✗ |

## A. On-Device Training

On-device training frameworks generally divide the learning process into three parts. *First*, the training framework must be able to detect when a significant shift has happened in the input dataset (**when to learn**). This can be done by calculating the per-output covariate distribution divergence on principal feature components [183], running mean and variance of streaming input [184], or confidence score of predictions [186]. *Second*, the on-device training framework must perform model adaptation within device constraints and limited training samples (**how to learn**). Three key techniques have been proposed for on-device model adaptation.

*1) Last Layer Transfer Learning:* The last layer of the network is fine-tuned through stochastic gradient descent one sample at a time [184] or reusing the outputs of feedforward execution without backpropagation [183] for batch gradient descent. Due to limited capacity and catastrophic forgetting, this approach results in poor performance when the distribution of new data is significantly different from the original training set [19].

*2) Train Specialized Operators:* TinyTL [19] proposed the use of lite residual learning modules for refining the output feature maps when updating just the bias instead of weights during on-device training to recoup performance loss. ML-MCU [185] proposed a lightweight one-versus-one (OVO) binary classifier for multiclass classification, which trains only those base classifiers that have a significant impact on final accuracy. This approach yields significant accuracy improvement over transfer learning (e.g., 34.1% higher than last layer transfer learning by TinyTL) without additional memory overhead but limits the application space due to constrained network types. Furthermore, TinyTL is not suitable for extremely resource-limited microcontrollers (e.g., Cortex-M).

*3) Special Learning Techniques:* Quantized continual learning prevents catastrophic forgetting by storing activation maps from past training data in the quantized form in a latent intermediate layer as replay data [188]. This allows learning from non-IID data. Incremental training uses constrained optimization to update the weights one sample at a time [186]. Both approaches suffer from limited application space due to limited supported network types. In addition, continual learning has high compute cost [188].

*Third*, the training framework must be able to select the samples to pick for training to maximize the learning effect, especially to prevent catastrophic forgetting for transfer learning approaches (**what to learn**). Common techniques include selecting samples based on their gradient norm, oversampling minority classes, and using weighted replay or sample importance weighing [183], [187]. Unfortunately, none of the on-device training frameworks is directly compatible with popular TinyML software suites, as none of the software suites is capable of unfreezing the frozen model graph on board. Moreover, all on-device training frameworks only work with networks having simple and limited architectural choices to prevent resource overflow. Thereby, additional memory constraints need to be injected into NAS frameworks to limit the model complexity.

## B. Federated Learning

FL extends on-device training to a distributed and non-IID setting, where the edge devices update parameters of a shared model on board, send the local versions of the updated model to a server, and receive a common and robust aggregated model, without the data ever leaving the edge devices [190]. We compare different FL frameworks suitable for TinyML listed in Table X using five distinguishing properties.

*1) FL Strategy:* The FL strategy refers to the selection of FL algorithms that the frameworks provide. Most FL frameworks provide a vanilla federated averaging (FedAvg) algorithm, where the local model weights are aggregated at the server instead of the gradients for communication efficiency [198]. Several enhancements to FedAvg have emerged to handle heterogeneity and resource constraints of AI-IoT devices. These include variants that have the following properties:

1) robust to laggards or client disconnections [190];
2) achieves similar accuracy across all devices [190];
3) includes device-specific model pruning to improve communication and training cost [192], [194];
4) uses transfer learning or fine-tuning for local model updates to save memory and build personalized models [195], [197];
5) uses knowledge distillation to aggregate class probabilities instead of weights [197].

Wu et al. [197] showed that transfer learning and knowledge distillation variants provide 5%–11% accuracy improvement over vanilla FedAvg for human action recognition while providing 10–5000× reduction in communication cost. Pang et al. [196] proposed the use of RL for model aggregation, obtaining 1.4%–2.7% higher accuracy over FedAvg for image recognition.

*2) Communication Stack:* FL requires a robust and efficient communication stack between the server and edge devices. Most FL frameworks rely on the robustness and efficiency guarantees provided by FedAvg and other FL strategies, such as the use of pruning or knowledge distillation over class probabilities [192], [194], [197]. Flower [190] and DIoT [193] provide bidirectional gRPC and WebSocket protocols to provide low-latency, concurrent, and asynchronous communication between server and clients. Both protocols are language, serialization, and communication agnostic.

*3) Scalability and Heterogeneity:* FL frameworks must be able to run workloads on hardware with different compute and communication budgets in a scalable fashion. *First*, the frameworks must be able to detect and track resource and task completion measures. Flower [190] includes a virtual client engine for scheduling and resource management. Fed-PARL [192] provides a resource and trust value tracker to monitor resource availability, bandwidth, task completion, task delay, and model integrity. DIoT [193] uses an unsupervised learning method to identify device state and type based on network traffic. *Second*, the frameworks should have a course of action for optimal workload distribution among these clients based on detected measures. The proposed techniques include partial work (average model weights based on gradient update sample count instead of timeout threshold) [190], [192], importance sampling (improve client selection probability of least-contributing clients) [190], adaptive pruning [194], and RL-based automated collaboration scheme discovery [196]. *Third*, the proposed techniques must generalize to a large number of clients. Among the different FL TinyML frameworks, Flower [190] has been shown to scale to 15M clients (1000 concurrent clients).

*4) Privacy:* FL frameworks must ensure that the local or global models cannot be reverse-engineered to uncover client data. Most FL frameworks for TinyML rely on the assumption that weight updates cannot be reverse-engineered to uncover local data. However, membership inference [203] and model inversion attacks [204] are successful against vanilla FedAvg. As a result, Flower [190] proposed using secure aggregation in their framework instead of vanilla model aggregation. The proposed semihonest protocol is robust against client dropouts, uses a multiparty computation protocol that does not require trusted hardware, and has low compute and communication overhead [205].

*5) Client Hardware and Supported Languages:* Finally, the FL frameworks must support a wide variety of clients with different processors and operating languages. Among the different frameworks, Flower [190], PruneFL [194], and TinyFedTL [195] were tested on microcontrollers, supporting Python, Java, and C++.

## X. KEY APPLICATIONS

Depending on the application, several variants of the TinyML workflow are used. In this section, we provide application-specific numerical insights from these workflows.

TABLE XI
SUMMARY OF IMAGE RECOGNITION FOR MICROCONTROLLERS

| Method | Dataset | Accuracy | SRAM (kB) | Flash (kB) | MACs (M) |
|---|---|---|---|---|---|
| ResNet8 [9] | CIFAR-10 | 85% | - | 96 | 25.3 |
| FastRNN [59] | Pixel MNIST-10 | 96% | <32 | 166 | - |
| FastGRNN [59] | | 98% | | 6 | |
| MCUNetV2-M4 [121] | ImageNet | T1- 65%, T5- 86% | 196 | 1010 | 119 |
| | Pascal VOC | mAP: 64.6% | 247 | <1000 | 172 |
| MCUNetV2-H7 [121] | ImageNet | T1- 72%, T5- 91% | 465 | 2032 | 256 |
| | Pascal VOC | mAP: 68.3% | 438 | <2000 | 343 |
| $\mu$NAS CNN [122] | CIFAR-10[B, M] | 77-86% | 0.9-15.4 | 0.7-11.4 | 0.04-0.38 |
| | MNIST | 99% | 0.49 | 0.48 | 0.029 |
| | Fashion MNIST | 93% | 12.6 | 63.6 | 4.4 |
| SpArSe CNN [86] | CIFAR-10[B, M] | 73-82% | 1.2 | 0.78 | |
| | MNIST | 97-99% | 1.3-1.9 | 1.4-2.8 | |
| | Chars74k[B] | 78% | 0.72 | 0.46 | |
| ProtoNN [58] | CIFAR-10[B] | 76% | | 15.9 | |
| | WARD[B] | 96% | | 15.9 | |
| | MNIST[B, M] | 96% | | 16-63.4 | |
| | USPS[B, M] | 95-96% | | 11.6-64 | |
| | CURET | 94% | - | 63.1 | - |
| Bonsai [57] | CIFAR-10[B] | 73% | | 0.5 | |
| | WARD[B] | 96% | | 0.47 | |
| | MNIST[B, M] | 94-97% | | 0.49-84 | |
| | USPS[B] | 94% | | 0.5 | |
| | CURET | 95% | | 115 | |
| | Chars74k[B,M] | 59-75% | | 0.5-101 | |
| SqueezeNet [92] | ImageNet | T1- 58%, T5- 80% | | 470-4800 | 349-848 [199] |
| Compressed LeNet [49] | MNIST | 98-99% | | 27-44 | - |
| AttendNets [114] | ImageNet | T1- 72-73% | | ∼1000 | 191-277 |
| AttendSeg[∧] [115] | CamVid | 90% | | 1190 | 7450 |
| ASL CNN [200] | Kaggle ASL | 75-99% | <400 | 185 | |
| Masked Face CNN [201] | Custom Masked Face | 99% | <400 | 128 | - |
| RaScaNet [202] | Pascal VOC[B] | 83-86 | 4-8 | 31-46 | 9.7-56.3 |
| RNNPool MbNetv2 [60] | ImageNet | T1- 70% | 240 | <2000 | 226 |
| Batteryless CNN [24] | MNIST | 99% | <8 | <256 | - |
| FOMO [32] | Beer and Can | 96% | 244 | 77.6 | |

B = binary dataset, M = multiclass dataset (assume M if unspecified).
∧ semantic segmentation from video.
mAP: mean average precision, T1: top 1%, T5: top 5%.

TABLE XII
SUMMARY OF VISUAL WAKE WORDS' DETECTION FOR MICROCONTROLLERS

| Method | Accuracy | SRAM (kB) | Flash (kB) | MACs (M) |
|---|---|---|---|---|
| MobileNetV1 [8] | 80% | - | 325 | 15.7 |
| MicroNets MbNetV2 [8] | 78-88% | 75-275 | 250-800 | - |
| MCUNetV2 [121] | 90-94% | 30-118 | <1000 | |
| RaScaNet [202] | 88-92% | 4-8 | 15-60 | 8-57 |
| RNNPool MbNetv2 [60] | 86-91% | 8-32 | 250 | 38-53 |
| MNasNet [14] | 85-90% | 50-250 | 400 | 10-54 |

Dataset: Visual Wake Words [14].

### A. Image Recognition and Visual Wake Words

Since the inception of AlexNet in 2012 [93], DNNs have been extensively used for visual understanding, such as image classification, object detection, handwriting recognition, visual wake words' detection, and semantic segmentation [14], [206]. The trend has trickled into the TinyML community as well, as evident in Tables XI and XII. Image recognition on the CIFAR-10 dataset and person detection on the Visual Wake Words dataset are two inference benchmarks in the MLPerf Tiny v0.5 [9]. Among the techniques shown in Table XI, NAS on residual convolutional architectures (e.g., MCUNetV2 [121], $\mu$NAS [122], and SpArSe [86]), rapid downsampling (e.g., RNNPool [60]), sparse projection (e.g., Bonsai [57] and ProtoNN [58]), and deep compression (e.g., Compressed LeNet [49] and SqueezeNet [92]) are the most common. Models that operate on multiclass datasets are suitable for Cortex M class architectures, while models that operate on binary datasets have been shown to be deployable on AVR RISC microcontrollers. In Table XI, MCUNetv2 [121] and AttendNets [114] achieved the state-of-the-art top 1% accuracy (72%–73%) on ImageNet for microcontrollers. MCUNetv2 uses once-for-all NAS on convolutional operators, combined with patch-by-patch inference and receptive field redistribution during runtime [121]. AttendNets use a standalone visual attention condenser, which improves spatial-channel selective attention [114]. MCUNetv2 [121] (with a YOLOv3 backbone) and AttendSeg [115] (with attention condensers) achieved state-of-the-art performance for semantic segmentation on the Pascal VOC and CamVid datasets, respectively. $\mu$NAS CNN achieved state-of-the-art performance on CIFAR-10 and MNIST [122]. Two interesting applications that deviate from traditional machine vision datasets include American sign language prediction [200] and detecting face masks in light of COVID-19 [201].

Detecting visual wake words (i.e., person detection) is a special case of image recognition. Table XII lists some of the models proposed for performing wake words' detection on the visual wake words dataset [14]. Among all the proposed models, RaScaNet [202] achieves the best balance of accuracy and resource usage. RaScaNet extracts features from an image patch using convolutional blocks and then sequentially learns the latent representation of the entire image using recurrent blocks. The network also includes both spatial attention and channel attention to focus on spatially distinct and multihead discriminative feature representations.

### B. Audio Keyword Spotting and Speech Recognition

Voice is a core component in human–computer interaction. Audio keyword spotting or wake word detection is used in voice assistants to identify waking keywords (e.g., "Hey Google,[1]" "Hey Siri,[1]" or "Alexa[1]"). The assistants must continuously listen for the keyword in utterances without being power- or resource-hungry [9]. Table XIII lists some keyword spotting, speech enhancement, and wake words' detection models geared toward microcontrollers. The use of lightweight depthwise-separable convolution, attention condensers, and recurrent units has generated models that are in the order of $10^0$ kB. Some of the models (e.g., FastGRNN, ShallowRNN, and ULP RNN) can even run on AVR RISC microcontrollers with 2-kB SRAM, while others are suitable for deployment on Cortex M class microcontrollers. The models typically operate on log Mel-spectrograms, which are short-time Fourier transforms transferred to the Mel scale [210] and available on the CMSIS-DSP library for embedded implementation. Most models use the Google Speech Commands dataset for training, which has 35 words with 105 829 utterances from 2618 speakers [10].

[1]Trademarked.

TABLE XIII
SUMMARY OF AUDIO KEYWORD SPOTTING AND SPEECH RECOGNITION FOR MICROCONTROLLERS

| Method | Dataset | Accuracy | SRAM (kB) | Flash (kB) | MACs (M) |
|---|---|---|---|---|---|
| DS-CNN [9] | SC* | 92% | - | 52.5 | 5.54 |
| MicroNets DS-CNN [8] | | 96% | 103 | 163 | 16.7 |
| FastRNN [59] | | 92% | < 2-32 | 56 | - |
| | STCI@ | 97% | | 8 | |
| FastGRNN [59] | SC | 92% | | 5.5 | |
| | STCI | 98% | | 1 | |
| ShallowRNN [169] | SC | 94% | 1.5 | 26.5 | 0.59 |
| | STCI | 99% | | | 0.3 |
| MCUNet DS-CNN [31] | SC | 96% | 311 | <1000 | - |
| $\mu$NAS CNN [122] | | 96% | 21 | 37 | 1.1 |
| Hello Edge DS-CNN [13] | | 94% | < 320 | 38.6 | 5.4 |
| TinySpeech-Z [113] | | 92% | | 21.6 | 2.6 |
| LMU-4 [207] | | 93% | - | 49 | - |
| Kronecker LSTM [90] | | 91% | | 8 | 0.02 |
| TinyLSTM^ [208] | CHiME2 | SDR: 13.0 dB | 3.7 | 310 | 0.66 |
| ULP RNN$^\vee$ [209] | SC, MUSAN | <3% NTR | - | 0.52 | - |

* SC refers to the Google Speech Commands dataset.
@ STCI refers to the Microsoft STCI Wake Words dataset.
^ for speech enhancement.
$\vee$ for wake-words detection in a noisy environment.

### C. Anomaly Detection

Anomaly detection or one-class classification detects outliers or deviations in the input data stream to indicate malfunctions [120] in an unsupervised fashion. Included in MLPerf Tiny v0.5 benchmark, applications of anomaly detection include diagnosis of industrial machinery [8], [9], [120], physiological disorders (e.g., heart attacks and seizures) [171], and climate conditions [211]. The two most common network architectures for microcontroller-based anomaly detection are fully connected autoencoders (FC-AEs) and depthwise CNN. Table XIV lists some of the anomaly detectors developed for microcontrollers. Among the different techniques, DROCC can operate directly on raw audio, sensor data, and images [171] without feature extraction. DROOC assumes that normal points lie on a low-dimensional linear manifold, while points surrounding the normal points outside a threshold radius are outliers, which can be augmented in a generative adversarial manner into the training set. Other audio-based anomaly detectors generally operate on mel-spectrograms [8], [9], [120].

### D. Activity and Gesture Tracking

Activity and gesture tracking form the central oracle for many applications, including health monitoring, behavioral analysis, context detection, augmented reality, and speech recognition [2]. Table XV showcases some activity detection framework geared toward microcontrollers. The common theme is to use lightweight models or use conventional models with a lower number of layers or polynomial complexity. Most models achieve accuracies of 90% or more for simple macrogestures (e.g., discrete fist gestures) or macroactivities (e.g., walking, running, standing, turning, and jumping) while being $10^0$ or $10^1$ kB order of size. The models are mostly hand-tuned due to the innate lightweight nature of the chosen models with a few automated using NAS.

TABLE XIV
SUMMARY OF ANOMALY DETECTORS FOR MICROCONTROLLERS

| Method | Dataset | Mean AUC | Peak SRAM (kB) | Flash (kB) | MACs (M) |
|---|---|---|---|---|---|
| OutlierNets [120] | MIMII | 0.83 | - | 2.7-26.7 | 2.87-22.9 |
| MicroNet DS-CNN [8] | ToyADMOS, MIMII | 0.96 | 114-383 | 253-442 | 38-129 |
| FC-AE [9] | ToyADMOS, MIMII | 0.85 | 4.7 | 270 | 0.52 |
| DROCC [171]* | CIFAR-10 | 0.74 | | 248 | 1.31 |
| | Thyroid | 0.78 | | 1.7 | 0.00031 |
| | Arrhythmia | 0.69 | - | 23.2 | 0.011 |
| | Abalone | 0.68 | | 1.9 | 0.00038 |
| | Epileptic Seizure | 0.98 | | 279 | |
| AnoML CNN [211] | AnoML | 0.57 | < 256 | 19.5-19.6 | - |

* DROCC uses an FC-AE for Thyroid, Arrhythmia, and Abalone datasets, LeNet-5 for CIFAR-10, and 1-layer LSTM for Epileptic seizure dataset.

TABLE XV
SUMMARY OF ACTIVITY AND GESTURE DETECTORS FOR MICROCONTROLLERS

| Method | Sensor(s) | Task | Accuracy | Flash |
|---|---|---|---|---|
| GesturePod ProtoNN [212] | MPU6050 inside white cane | Detect 5 cane gestures | 92% | 6 kB |
| AURITUS FastRNN [2] | eSense earable | Detect 9 macro activities | 98% | 6 kB |
| Bian et al. 1D-CNN [213] | Wrist-worn capacitive array | Detect 7 hand gestures | 96% | 30 kB |
| T'Jonck et al. CNN [214] | BMA400 inside mattress | Detect 5 bed activities | 89% | < 1 MB |
| Zhou et al. HDC + SVM [215] | MPU-6050 and EMG pad (wrist-worn) | Detect 13 hand gestures across 8 limb positions | 93% | 135 kB |
| Elsts et al. CNN [216] | Colibri Wireless IMU | Detect 18 macro activities | 73% (F1 score) | 20 kB |
| Coelho et al. DT [217] | Chest, waist and ankle-mounted IMU | Detect 12 macro activities | 97% | 22 kB |
| FastGRNN (LSQ) [57] | IMU on torso and limbs | Detect 6 macro activities and 19 sports activities | 96% 84% | 3 kB 3.25 kB |

### E. Odometry and Navigation

Odometry is the fusion of onboard sensors for indirect estimation of an object's position and attitude in the absence or conjunction with infrastructure-dependent localization services [218]. TinyOdom [1] exploits THIN-Bayes, TCN backbone, a magnetometer, physics, and velocity-centric sequence learning formulation to yield neural inertial odometry models that are 31–134× smaller than existing neural inertial odometry models, suitable for deployment on Cortex-M architectures. Vehicle neural networks (VNNs) [21] use a modified and quantized LeNet-5 as an autonomous controller on a car under stochastic lighting conditions. The network leverages imitation learning via a classical computer vision teacher algorithm for training. VNNs have 7.5–163.5 kMACs, and the PULP implementation on GAP8 SoC reduces latency and energy consumption by 13× (0.2–1.2 mS) and 3.2× (3.9–18.9 $\mu$J per inference), respectively, over Cortex-M architectures, achieving 97% accuracy. A class of residual networks intended for standard-UAV navigation without SLAM called DroNets [219] have been ported on nano-UAVs retrofitted with a PULP GAP8 SoC shield [220]. By using tiling, quantization, parallelization, and signal processing on the PULP chip, the platform achieved 6–18 FPS within a 64-mW power envelope, covering 113 m unseen indoor trajectory in the real world at

TABLE XVI
EXAMPLE TINYML MHEALTH APPLICATIONS FOR MICROCONTROLLERS

| Method | Task | Performance | Flash |
|---|---|---|---|
| TinyEats [221] | Dietary monitoring | Accuracy 95% | 35 kB |
| Arlene et al. [222] | Sleep apnea detection | Accuracy: 99% | 212 kB |
| Petrović et al. [223] | Cough detection | Accuracy 95% | 35 kB |
| DROCC [171]* | Detect cardiac arrythmia | AUC: 0.69 | 23.2 kB |
| | Detect epileptic seizure | AUC: 0.98 | 279 kB |
| | Detect hypothyroid condition | AUC: 0.78 | 1.7 kB |
| AURITUS Bonsai [2] | Earable fall detection | Accuracy: 98% | 2.3 kB |

TABLE XVII
SUMMARY OF FACE DETECTORS FOR MICROCONTROLLERS

| Method | Dataset | Performance | Peak SRAM | MACs |
|---|---|---|---|---|
| FaceReID [224] | VGGFace2 | Accuracy: 0.97 | 352 kB | 85M |
| Batteryless Face Detection [25] | CelebA | Accuracy: 0.97 | 384 kB | - |
| EagleEye [225] | WIDER FACE | mAP*: 0.77 | 1170 kB | 80M |
| RNNPool S3FD [60] | WIDER FACE | mAP*: 0.83 | 225 kB | 120M |
| MCUNetv2 S3FD [121] | WIDER FACE | mAP*: 0.89 | 672 kB | 110M |

* mean average precision for ≤3 faces

a speed of 1.5 m/s [22]. In all cases, the odometry models enjoy the lightweightness of classical odometry techniques and the resolution of large networks.

### F. mHealth

TinyML opens up a broad spectrum of real-time and low-footprint eHealth applications, some of which are summarized in Table XVI. These include monitoring eating episodes and coughs using microphones [221], [223], sleep monitoring and arrhythmia detection through ECG measurements [171], [222], epileptic seizure recognition from EEG sensors [171], and fall detection using earable inertial sensors [2]. Most TinyML mHealth applications are variants of anomaly detection, indicating the presence or absence of a health condition, thereby allowing the use of ultralightweight models in the order of $10^0$–$10^1$ kB. Example models for mHealth include Bonsai [2], embedded GRU [221], 1-D CNN [222], FC-AE [171], and two-layer CNN/LSTM [223].

### G. Facial Biometrics

Facial biometrics has been a prominent authentication technique in civilian and military applications [226]. Existing face detectors use deep learning to automate the feature representation pipeline while approaching human performance [226]. Table XVII illustrates some face detectors for microcontrollers. A common recipe for porting deep face detectors onto microcontrollers includes the use of lightweight spatial convolution coupled with NAS, quantization, and inference engine optimizations. Typical neural blocks include squeeze and excitation modules [224], coupling depthwise with pointwise convolution [225], and nonlinear pooling between convolutional layers [60]. Successive and rapid downsampling helps cut out redundant network layers further [60], [225] while ensuring scale-equitable face detection. Inference engine optimizations include patch-based inference scheduling [121],

TABLE XVIII
IMPACT OF FEATURE PROJECTION VERSUS RAW DATA INFERENCE

| Application | Method | Accuracy | Latency/MAC | Flash |
|---|---|---|---|---|
| Gesture Recognition[∨] | CNN [32] | 100% | 11 mS | 45.3 kB |
| | MLP [32] | 100% | 5 mS | 17.8 kB |
| Human Activity Recognition[@] | TCN [2] | 94.6% | 7.52M | 52.8kB |
| | FastGRNN [2] | 97.6% | - | 13.1 kB |
| | Bonsai [2] | 80.3% | 0.0136M | 14.8 kB |
| Anomaly Detection* | DROCC [171] | F1 - 68% | 0.00038M | 1.9kB |
| | OC-SVM [171] | F1 - 48% | - | 2.99 MB |

[∨]: Device: Cortex M7
*  Dataset: Abalone, [@] Dataset: AURITUS
■ Models operating on features

TABLE XIX
IMPACT OF COMPRESSION VERSUS NO COMPRESSION

| Application | Dataset | Method | Accuracy | Latency/MAC | Flash |
|---|---|---|---|---|---|
| Human Activity Recognition | HAR-2[∨] | FastGRNN [59] | 94.5% | - | 29 kB |
| | | FastGRNN-L [59] | 96.8% | - | 28 kB |
| | | FastGRNN-LS [59] | 96.3% | 172 mS | 17 kB |
| | | FastGRNN-LSQ [59] | 95.6% | 62 mS | 3 kB |
| | HAR-1[@] | BiLSTM [228] | 91.9% | 470 mS | 1.5 MB |
| | | BiLSTM-Prune [90] | 83% | 98.2 mS | 76 kB |
| | | BiLSTM-Q [90] | 91.1% | - | 384 kB |
| | | BiLSTM-KP [90] | 91.1% | 157 mS | 75 kB |
| Audio Keyword Spotting | Speech Commands* | FastGRNN [59] | 93.2% | - | 57 kB |
| | | FastGRNN-L [59] | 93.8% | - | 41 kB |
| | | FastGRNN-LS [59] | 92.6% | 779 mS | 22 kB |
| | | FastGRNN-LSQ [59] | 92.2% | 242 mS | 5.5kB |
| | | LSTM [13] | 92.5% | 26.8 mS | 243 kB |
| | | LSTM-Prune [90] | 84.9% | 5.9mS | 16 kB |
| | | LSTM-Q [90] | 92.2% | - | 65 kB |
| | | LSTM-KP [90] | 91.2% | 17.5 mS | 15 kB |
| Image Recognition | MNIST-10[@] | Bonsai [57] | 97% | - | 84 kB |
| | | Bonsai-Q [57] | ∼97% | - | 21 kB |
| | | LSTM [90] | 99.4% | 6.3 mS | 45 kB |
| | | LSTM-Prune [90] | 96.5% | 0.7 mS | 4.2 kB |
| | | LSTM-KP [90] | 98.4% | 4.6 mS | 4.1 kB |
| | ImageNet | SqueezeNet [92] | T1-57.5% | 848M | 4.8 MB |
| | | SqueezeNet-PQE [92] | T1-57.5% | 349M | 0.5 MB |
| | | AlexNet [57] | T1-57.2% | 723M | 240 MB |
| | | AlexNet-Prune [57] | 57.2% | - | 27 MB |
| | | AlexNet-PQ [57] | 57.2% | - | 9 MB |
| | | AlexNet-PQE [57] | 57.2% | ∼348M | 6.9 MB |

[∨] Device: SAM3X8E Cortex-M3, [@] Hikey 960 Cortex A73
* Device: SAM3X8E Cortex-M3 for the first four, Hikey 960 Cortex A73 for the last four
Q - Quantized, S - Sparsified, L - Low Rank Factorization,
KP - Kronecker Products, E - Huffman Encoding
■ Uncompressed and lightweight model, ■ Uncompressed and vanilla model,
■ Compressed and lightweight model, ■ Compressed and vanilla model receptive field redistribution [121], and dual-memory management [224]. Patch-by-patch inference allows operation on only a tiny region of the activation map, while receptive field redistribution shifts receptive field and FLOPs to later stages to mitigate peak memory usage and overlapping patches [121]. Dual-memory management swaps variables between flash and RAM whenever required [224].

## XI. DISCUSSION AND CASE STUDIES

In this section, we break down the end-to-end workflow and provide quantitative analysis of individual aspects of the workflow based on select examples from Sections IV to X. We discuss how individual aspects contribute to the overall execution and also describe qualitatively how individual techniques for one aspect impact the choices for other aspects.

### A. Feature Projection Versus No Feature Projection

Feature projection allows a domain expert to retain data variance while reducing data dimensionality [2]. Intuitively, this reduces the model complexity needed to capture the variations in input data, i.e., feature projection is useful for simplifying the architecture of non-TinyML models. Consider the gesture recognition example in Table XVIII. Both CNN and MLP achieve the same accuracy. However, the MLP pipeline, operating on spectral features, requires 2.5× less flash, runs 2.2× faster, and requires 18× less SRAM than the CNN pipeline, which operates on raw data. By leveraging domain knowledge, simpler models can achieve the same accuracy yet save memory and inference costs over complex models. Well-designed features (e.g., audio MFCC, spectrograms, and signal power) are also able to exploit DSP functions (e.g., CMSIS-DSP) and accelerators embedded in most microcontrollers [6]. However, if the extracted features are not sufficiently discriminatory due to a lack of domain knowledge, then the performance of models will degrade [227]. Consider the human activity recognition example in Table XVIII. Bonsai operates on five statistical features surrounding the signal amplitude, which are unable to sufficiently distinguish among activity primitives that are statistically similar (e.g., sitting and sleeping). Thus, Bonsai suffers a 17% accuracy drop while being similar in size to FastGRNN. To achieve the performance of complex models that do not operate on features while having the computational efficiency of simple models operating on handcrafted features, ultralightweight ML blocks are used. These blocks can often be much more efficient and accurate than models tied to a feature extraction pipeline, as poorly designed features can yield significant compute overhead [2]. For example, in Table XVIII, DROCC outperforms one-class SVM for anomaly detection not only in accuracy (+20% gain) but also in model size (1600× reduction). Unfortunately, the problem with lightweight models is twofold. *First,* most of these models do not have enough parameters to model globally significant features, failing to generalize to new data distributions in the field [2]. *Second,* most NAS frameworks, TinyML software suites, intermittent computing tools, and online learning frameworks lack support for deploying some of these models on commodity microcontrollers. Therefore, adopting feature extraction requires a careful understanding of the application constraints, striking a balance between the availability of domain knowledge, feasible model architecture sets, feature acceleration support, and support from model optimization and architecture search tools.

### B. Compression Versus No Compression

Exploiting sparsity and reducing bitwidth of models depend on three key factors.

*1) Large Sparse Versus Small Dense:* Large-sparse models (compressed and vanilla models) are known to outperform small-dense models (uncompressed and lightweight models) for a broad range of network architectures [48] in terms of compression ratio for comparable accuracy. This is evident from the speech commands and MNIST-10 examples in Table XIX. LSTM-Prune and LSTM-KP outperform Fast-GRNN and Bonsai, providing on average 12× model size reduction with only 2.3% accuracy loss. Moreover, on average, all uncompressed and vanilla models provided a 13.5× reduction in model size when pruned compared to 2.1× for lightweight models (FastGRNN). Therefore, sparsification is useful when working with vanilla models rather than lightweight ML blocks.

*2) Pruning and Quantization Gains:* Unstructured pruning and posttraining quantization offer performance gain in different dimensions. Generally, both pruning and quantization are applied jointly [49].

1) *Flash Savings:* Pruning is more aggressive in reducing the model size than quantization [49]. In Table XIX, on average, pruning provides 13.6× compression factor compared to 3.9× compress factor provided by quantization. Pruning combined with quantization provides a 16× reduction in model size on average.
2) *SRAM Savings:* Pruning is less likely to reduce working memory footprint than quantization. After pruning, the microcontroller still has to perform multiplication in the original floating-point bitwidth, whereas, in quantization, the bitwidth of the multiplication decreases.
3) Intuitively, pruning is less likely to reduce the inference latency compared to integer quantization in microcontrollers. The gains from the loss of redundant weights are lower than the gains from the integer matrix multiplication. Moreover, unstructured pruning can add processing and execution time overhead [85].
4) *Accuracy Loss:* Pruning often causes higher accuracy loss than quantization. In Table XIX, on average, pruning reduced accuracy by 4.9%, compared 0.4% from quantization. This is due to a higher degree of information loss in pruning as in quantization only the bitwidth is reduced.

*3) Support From HW/SW:* Not all microcontrollers and TinyML software suites support or can reap the benefits of quantization of intermediate or sub-byte bitwidth [81]. For example, TFLM does not support arbitrary bitwidth of weights and activations [55]. Most microcontrollers are limited by their SIMD bitwidth, unable to exploit low precision representation of neural networks fully [81]. Therefore, care must be taken to ensure that the chosen quantization scheme is compatible with the choice of microcontroller and TinyML software suites.

### C. Lightweight Models Versus Vanilla Models

Most model compression techniques cannot reduce the size of pretrained models without significant loss in accuracy (e.g., pruning and quantization result in 19× reduction in model size on average in Table XIX). In some cases, the pretrained model is too big to apply model compression feasibly for a microcontroller (e.g., in Table XIX, AlexNet can be reduced to 6.9 MB from 240 MB), or the pretrained model may not even be a neural network (e.g., in Table XX, SVM, Coarse DT, AdaBoost, and kNN are nonneural models). In such cases, lightweight ML blocks are adopted to reduce the model size and inference latency while maintaining or exceeding the accuracy of vanilla models. In fact, from Table XX, we see that lightweight ML blocks are commonly adopted when

TABLE XX
IMPACT OF LIGHTWEIGHT VERSUS VANILLA MODEL USAGE

| Application | Dataset | Method | Accuracy | Latency/MAC | Flash |
|---|---|---|---|---|---|
| Human Activity Recognition | HAR-2[∨] | FastRNN [59] | 94.5% | <172 mS | 29 kB |
| | | FastGRNN [59] | 95.4% | 172 mS | 29 kB |
| | | RNN [59] | 91.3% | 590 mS | 29 kB |
| | | LSTM [59] | 93.7% | OOM[∧] | 74 kB |
| | AURITUS* | TCN [2] | 95.12% | 1380 mS | 60 kB |
| | | SVM [2] | 99.9% | OOM | 23 MB |
| | | MLP [2] | 99.8% | OOM | 418 kB |
| | | Coarse DT [2] | 98.5% | OOM | 1100 kB |
| | | AdaBoost [2] | 98.7% | OOM | 81.6 MB |
| Audio Keyword Spotting | Speech Commands | DS-CNN [9] | 92% | 5.54M | 52.5 kB |
| | | TinySpeech-Z [113] | 92.4% | 2.6M | 21.6 kB |
| | | LMU-4 [207] | 92.7% | - | 49 kB |
| | | CNN [229] | 90.7% | 76M | 556 kB |
| Image Recognition | MNIST-10 | Bonsai [57] | 97% | - | 84 kB |
| | | ProtoNN [58] | 95.9% | - | 63 kB |
| | | kNN [57] | 94.3% | OOM | 184 MB |
| | | MLP [57] | 98.3% | OOM | 3.1 MB |
| | ImageNet | AttendNets [114] | T1-71.7% | 191M | ∼1 MB |
| | | SqueezeNet[@] [92] | T1-57.5% | 848M | 4.8 MB |
| | | AlexNet [93] | T1-57.2% | 723M | 240 MB |

[∨] Device: SAM3X8E Cortex-M3, * Device: STM32 Cortex-M4 and M7
[∧] OOM = Out of memory on tested microcontrollers
[@] Uncompressed
▇ Vanilla models

out-of-memory errors are encountered on the microcontroller. For the human activity recognition (AURITUS) and image recognition (MNIST-10 and ImageNet) use cases, the vanilla models (SVM, MLP, Coarse DT, AdaBoost, and AlexNet) were simply too big to run on commodity microcontrollers, forcing the adoption of lightweight ML operators (Bonsai, ProtoNN, TCN, AttendNets, and SqueezeNet). In some cases, lightweight models are adopted to improve the accuracy and latency (e.g., FastGRNN has higher accuracy and lower latency than RNN). However, special attention must be paid to the specific compute budget when adopting these lightweight models. *First,* some of these models might improve the metrics in one dimension and degrade other dimensions. For example, in Tables XIX and XX, SqueezeNet has lower model size but higher latency (and energy usage) than AlexNet [230]. *Second,* as discussed earlier in the feature projection case study, some lightweight models overfit the training set and fail to generalize to unseen data. For example, in Table XX, TCN has a 5% reduction in test accuracy over SVM, MLP, coarse DT, and AdaBoost. In fact, for activity detection, Saha et al. [2] showed that lightweight ML blocks have an accuracy drop of 11.8% for the same test set distribution shift over vanilla models. *Third,* not all aspects of the TinyML workflow support every lightweight ML block. For example, $\mu$NAS [122], MicroNets [8], and SpArSe [86] assume a CNN backbone, while Sklearn Porter [174] only supports porting MLP to microcontrollers. Moreover, most on-device learning frameworks only support CNN backbones. Thus, the choice of lightweight ML blocks is limited by what the other components in the TinyML workflow support.

### D. Using NAS Versus Handcrafted Models

NAS is used when one or more model performance metrics (e.g., latency, SRAM usage, and energy) need to be

TABLE XXI
IMPACT OF NAS VERSUS HANDCRAFTED MODELS

| Application | Dataset | Method | Accuracy | Latency/ MAC | Flash |
|---|---|---|---|---|---|
| Inertial Odometry[○] | OxIOD | L-IONet TCN [231] | 2.82 m | 13.9M | 183 kB |
| | | RoNiN TCN [232] | 0.42m | 220M | 2.1 MB |
| | | TinyOdom TCN [59] | 1.24m-1.37 m | 4.64M-8.92M | 71 kB-118 kB |
| Audio Keyword Spotting | Speech Commands | DS-CNN [9] | 92% | 5.54M | 52.5 kB |
| | | MicroNets DS-CNN [8] | 95.3% - 96.5% | 16M-129M | 102 kB-612 kB |
| | | $\mu$NAS-CNN [122] | 95.4-95.6% | 1.1M | 19 kB-37 kB |
| Image Recognition | Visual Wake Words[*] | MBNetv2 [8] | 86% | 0.46s | 375 kB |
| | | MicroNets MBNetv2 [8] | 78.1%-88% | 0.08s-1.13s | 230 kB-833 kB |
| | MNIST-10[*] | Bonsai [57] | 94.4% | 8.9 mS | 1.97 kB |
| | | SpArSe-CNN [86] | 95.8%-97% | 27mS-286mS | 2.4kB-15.9kB |
| | CIFAR-10[B,*] | Bonsai [57] | 73% | 8.2 mS | 1.98 kB |
| | | SpArSe-CNN [86] | 70.5%-73.4% | 0.49s-2.52s | 2.7 kB-9.9 kB |
| | | $\mu$NAS-CNN [122] | 77.5% | - | 0.69kB |

[○] Accuracy metric is relative trajectory error [232] (lower is better)
[*] Device: STM32 Cortex-M4 and M7, [B] Binary dataset
■ Handcrafted models

TABLE XXII
IMPACT OF RUNTIME OPTIMIZATIONS VERSUS NO OPTIMIZATIONS

| Application | Dataset | Method | Accuracy | Latency/ MAC | Flash |
|---|---|---|---|---|---|
| Human Activity Recognition | Custom[*] | CNN-TFLM [189] | 85% | 58 mS | 275 kB |
| | | CNN-Cube.AI [189] | 85% | 14 mS | 192 kB |
| Audio Keyword Spotting | Speech Commands[*] | CNN-TFLM [189] | - | 380 mS | 288 kB |
| | | CNN-Cube.AI [189] | - | 373 mS | 247 kB |
| Image Recognition | ImageNet | MCUNet MbNetv2 [121] | 60.3%-68.5% | 68M-126M | 1MB-2MB |
| | | MCUNetV2 MbNetv2 [121] | 64.9%-71.8% | 119M-256M | 1MB-2MB |
| | Pascal VOC | MbNetv2+CMSIS [121] | mAP: 31.6% | 34M | OOS |
| | | MCUNetV MbNetv2 [121] | mAP: 51.4% | 168M | <2 MB |
| | | MCUNetV2 MbNetv2 [121] | mAP: 64.6% | 172M | <1 MB |
| | CIFAR-10[@] | CNN [164] | 80.3% | 456 mS | < 1 MB |
| | | CNN-CMSIS [164] | 80.3% | 99 mS | < 1 MB |

[*] Device: STM32 Cortex-M4, [@] Device: STM32 Cortex-M7, OOS: Overflowed SRAM
■ Superior optimization techniques than comparing method in the same dataset class constrained to suit the deployment scenario. NAS is particularly useful in three cases.

*1) Metrics Form Competing Objectives:* The most common motivation behind NAS is to increase the model accuracy while decreasing the flash, SRAM, latency, and energy usage. These metrics form competing objectives under search space and device constraints. For example, a larger model is likely to provide higher accuracy but consume more flash and SRAM. A certain architecture (e.g., SqueezeNet) is likely to reduce flash usage but can have higher latency than a larger model (e.g., AlexNet). The model might have to follow certain bounds or rules (e.g., cannot use a specific operator type). In Table XXI, the goal is to find the best performing models that reach the desired objectives within the specified constraints. In all cases, the NAS strategy consistently outperforms handcrafted models in terms of providing the most accurate model within the device constraints.

*2) Optimize High-Dimensional Search Space for Multiple Target Hardware:* Neural network search spaces can grow intractable quickly. For example, the search space of a CNN can contain the number of layers, the number of kernels in each layer, the size of the kernel in each layer, the stride in each layer, the size of kernels in the pooling layer, and so on [122]. The search space might even contain parameters for different model architectures. Furthermore, the network might have to be optimized for multiple microcontrollers with distinct compute and memory budget [5]. To save human time and effort, NAS algorithms can automatically perform model architectural adaption to fully exploit the target capabilities of different hardware. In the inertial odometry example in Table XXI, TinyOdom [1] produces four different models that provide a competitive resolution within the memory constraints of four different microcontrollers, providing a 1.6–30× reduction in model size while suffering a resolution drop of 1.2× compared to handcrafted models. Similarly, in the keyword spotting example in Table XXI, MicroNets [8] generates three different DS-CNNs that are suitable for three different microcontroller models, outperforming the handcrafted DS-CNN by 3.3–4.5%.

*3) Prior Wisdom Does Not Suit Deployment Needs:* In some deployment scenarios, expert knowledge may not suit the deployment needs. For example, in the inertial odometry case in Table XXI, TinyOdom [1] was the first framework allowing the deployment of inertial odometry models on microcontrollers. In the case of image recognition, $\mu$NAS [122] attempted to deploy the models on AVR RISC microcontrollers, which have a much tighter resource budget than Cortex M4 microcontrollers used by Bonsai [57] and SpArSe [86]. Similarly, SpArSe [86] attempted to run DNNs on microcontrollers and not nonneural models to broaden the application spectrum of AI-IoT. Under unexplored circumstances, NAS can bring valuable insights during model discovery on achievable performance and optimal architectural choices.

### E. Using Runtime Optimizations Versus No Optimizations

The use of TinyML software suites to generate code and perform operator/inference engine optimizations is a mandatory step in the TinyML workflow, often needed to guarantee the execution of a trained model on the microcontroller. Consider the CIFAR-10 image recognition example in Table XXII. The use of partial *im2col* in CMSIS allows the CNN to have a working memory of 133 kB instead of 332 kB, in which case the CNN would overflow the Cortex-M7 SRAM [164]. The optimized operator set also reduces the inference latency by 4.6× and decreases energy usage by 4.9× [164]. Similarly, MCUNetv2 [121] achieved record ImageNet and Pascal VOC accuracy on microcontrollers by optimizing a large MBNetv2 that normally overflows the SRAM using patch-by-patch inference and receptive field redistribution. However, to pick the appropriate software suite, other questions must be asked.

1) Which microcontrollers are suitable for my application?
2) What are the memory, latency, and energy requirements?
3) Which ML blocks are suitable for my application?
4) Which training frameworks can I use?
5) Do I need support for intermittent computing?
6) Do I need support for online learning?
7) Do I need an automated schedule explorer?
8) Is dynamic memory management necessary?
9) How many models need to run on the same platform?
10) Do I need to share the same model across platforms?
11) Do I need to sparsify or quantize any model?

Consider the human activity recognition and keyword spotting use case in Table XXII. TFLM uses an interpreter-based approach to realize the model graph during runtime [167]. TFLM supports dynamic memory management {7)}, multitenancy {8)}, and updating the model binaries rather than the entire codebase for fast prototyping and portability across platforms {9)}. However, {7)}–{9)} come at 1.3× increase in flash usage and 2.6× increase in latency (lagging on {2)}) compared to STM32Cube.AI, which embeds operator function calls into native C code [172]. STM32Cube.AI, on the other hand, only supports STM32 series of Cortex-M microcontrollers (lagging on {1)}), while TFLM provides much broader platform support. However, STM32Cube.AI also supports nonneural model deployment (e.g., k-means, SVM, RF, kNN, and DT), while TFLM only supports neural network deployment (lagging on {3)}). Likewise, {5)} can only be realized through SONIC and TAILS [24], and {7)} is provided by only microTVM [130]. Both quantitative and qualitative tradeoffs similar to the case study here must be performed to pick the appropriate software suite.

TABLE XXIII
IMPACT OF ONLINE LEARNING VERSUS STATIC MODELS

| Application | Dataset | Method | Accuracy | Latency |
|---|---|---|---|---|
| Image Recognition* | MNIST-10 | CNN [183] | 10%-82% | - |
| | | CNN (LW) [183] | 65%-98% | |
| | CIFAR-10 | CNN [183] | 12%-38% | |
| | | CNN (LW) [183] | 55%-68% | |
| Anomaly Detection @ | Custom | Autoencoder [184] | 75% | 1.75 mS |
| | | Autoencoder (TinyOL) [184] | 100% | 1.92 mS |

* Device: TI MSP430
@ Signal reconstruction error, Device: nRF52840 Cortex-M
■ No online learning

### F. Using Online Learning Versus Static Models

Online learning improves the performance of the model by adapting the model on board without sensitive data leaving the device. Consider the case studies on online learning shown in Table XXIII. The performance of models improves by 34% when on-device training is used to adapt to dataset shifts. For TinyOL, the latency overhead to include online learning is 10%. While the performance gains are somewhat transparent, the major barrier in on-device learning is the lack of support from other aspects of the workflow. For example, most on-device learning frameworks assume the use of CNN or binary classifiers and also use a custom code generator for the model due to a lack of online learning support from existing TinyML software suites. Moreover, it is not clear how NAS should account for the training memory and inference overheads when on-device learning is used. The lack of comprehensive studies of on-device learning also limits the adoption of FL in TinyML. Particularly, while the TinyML workflow was designed for a single noncollaborative model, FL requires the distribution of a global model to be enhanced via local model updates. While existing FL frameworks have tools to distribute resources heterogeneously, it is unclear how NAS, model compression, or lightweight ML blocks affect the real-world setting, as none of the FL frameworks have studied these effects. Thereby, online learning constrains the user to a very specific choice of models and custom software suites.

## XII. CHALLENGES AND OPPORTUNITIES

The first-generation efforts in TinyML focused on the engineering and mechanics of squeezing ML models within the limited memory, compute, and power bounds of a microcontroller. Both academia and industry have established several TinyML software frameworks to streamline the deployment of ML models for microcontrollers. Many of the issues raised by prior surveys [3], [4], [6] have been addressed. However, the following new challenges are emerging that require further research.

### A. Application Specific Safety and Heuristic Requirements

Real-world IoT applications operate within certain bounds, correlations, and heuristic rules set forth by the operating domain and system physics. For example, a UAV cannot exceed a certain bank angle without compromising stability [233]. In complex event processing, specific granular action primitives (e.g., cooking a dish) must always precede other primitives (e.g., chopping vegetables) [234]. Neural networks cannot assure that the learned distributions obey all the laws [235]. As a result, recent neural network pipelines are being injected with trainable neurosymbolic reasoning [236], [237], signal temporal logic [235], and physics-aware embeddings [238], [239], [240], [241] for robust complex event processing within the laws and bounds of physics. For making rich and complex inferences beyond binary classification, the TinyML workflow requires research to combine data and human knowledge by including logical reasoning modules within the microcontroller's compute and memory bounds.

### B. Data Quality and Uncertainty Awareness

Sensor data in the wild suffer from missing data, cross-channel timestamp misalignment, and window jitter [227], [242]. These uncertainties may stem from scheduling and timing stack delays, system clock imperfections, sensor malfunction, memory overflow, or power constraints [243], [244]. Sensing uncertainty can reduce the performance of ML models when training for complex event processing [227]. TinyML models need to be injected with uncertainty awareness by incorporating appropriate training frameworks [227], [242] in the workflow or use onboard clocks and hardware enhancements for precise time synchronization [245].

### C. On-Device Fine-Tuning

Models in the wild need to be fine-tuned periodically to ensure robustness across domain shifts in incoming data distribution [183]. *First*, while several on-device learning frameworks have been proposed for edge devices [246], they either work on high-end edge devices (e.g., Raspberry Pi) [19], [247] or can update weights of a few layers on microcontrollers [183]. Software-centric resource constraints, constrained learning theories, and static resource budget prevent on-device learning from being a viable alternative to cloud-based training for microcontrollers [246]. *Second*, an alternate line of work suggests low-latency compressive offloading onto the cloud [20] but has nondeterministic compression ratios and offloading points. *Finally*, the models themselves can be made more robust to domain shifts through representation learning [248] or domain-adversarial training [249], but the resulting models do not fit on microcontrollers. More work needs to be done in striking an optimal balance between on-device fine-tuning and over-the-air model updates, and whether unsupervised embeddings can be ported onto microcontrollers.

### D. Backward Compatibility

The changes in behavior when deploying an upstream model (e.g., a model on the cloud) to microcontrollers through the TinyML workflow cannot be measured in isolation using only the aggregate performance measures (such as accuracy) [250]. Even when a TinyML model (downstream model) and the upstream model have the same accuracy, they may not be functionally equivalent and may have samplewise inconsistencies [251] resulting in new failures impacting high-stake domains, such as healthcare. This notion of functional equivalence between an upstream and a downstream model is known as backward compatibility. When previously unseen errors are observed in the downstream model, the downstream model is said to be backward incompatible [252] and has low fidelity [253] and high perceived regression [251] with respect to the upstream model. As a result, to have robust inference, the TinyML model must have both high accuracy and high fidelity with its upstream counterpart. The proposed solutions, such as positive congruent training [251] and backward compatible learning [254], are yet to be integrated and optimized for the TinyML workflow.

### E. New Security and Privacy Threats

While constraining private data within the IoT node reduces the chance of privacy and security leaks associated with cloud-based inference, the attack surface on TinyML platforms is wide open. Compressed models are prone to adversarial attacks and false data injection with a higher success rate than larger models [255], [256], [257]. At the sensing layer, microarchitectural and physical side channels can leak information from microcontroller chips through cache leaks, power analysis, and electromagnetic analysis [258]. Direct attacks on IoT devices include malware injection, model extraction, access control, man-in-the-middle, flooding, and routing [258]. Therefore, the NAS optimization function in the TinyML workflow should include adversarial robustness goals to provide not only the smallest models but also the models most robust to adversarial attacks [256], [257], [259]. The workflow should also include attack surface analysis and tools to defend the inference pipeline against attacks.

### F. Hardware/Software Coexploration

Much of the development in TinyML has been software-driven, with the hardware platform being static. While IoT platforms hosting microcontrollers are shrinking due to Moore's law, the workload and the complexity of neural networks have skyrocketed [7], [260]. The proposed hardware innovations include the use of a systolic array, stochastic computing, in-memory computing, near-data processing, spiking neural hardware, and non-von Neumann architectures [7], [260], [261]. However, such architecture innovations are largely disjoint from the TinyML software communities. Developments in TinyML software need to be performed hand-in-hand with attention-directed hardware design with the platform and model being optimized jointly [262], [263].

## XIII. Conclusion

It is desirable to enable onboard ML on microcontrollers, turning them from simple data harvesters to learning-enabled inference generators. To that end, we introduced a widely applicable workflow of ML model development and deployment on microcontroller-class devices. Several applications are showcased to highlight the tradeoffs in different instances of this workflow adoption. Although the current efforts can transition the state-of-the-art ML models to ultraresource-constrained environments, we consider them as the first generation of TinyML and present new opportunities. Through this review, we envision a need for the next generation of TinyML frameworks to address the discussed challenges that have received limited explorations.

**Swapnil Sayan Saha** (Graduate Student Member, IEEE) received the B.Sc. degree in electrical and electronic engineering (EEE) from the University of Dhaka, Dhaka, Bangladesh, in 2019, and the M.S. degree in electrical and computer engineering (ECE) from the University of California at Los Angeles (UCLA), Los Angeles, CA, USA, in 2021, where he is currently pursuing the Ph.D. degree in ECE with the Networked and Embedded Systems Laboratory.

To date, he has published more than 20 peer-reviewed conference articles, journal articles, and book chapters. His research explores the creation of deployable, learning-enabled, and resource-constrained sensing systems.

Mr. Saha received more than 30 awards in robotics, technical, and business-case competitions worldwide, including the IEEE Richard E. Merwin Scholarship, the IEEE Lance Stafford Larson Best Paper Award, the All IEEE Young Engineers' Humanitarian Challenge, the Rise High Bangladesh, and the EMK Center Science for Mankind Research Award.

**Sandeep Singh Sandha** received the B.Sc. degree in computer science (CS) from IIT Roorkee, Roorkee, India, in 2014, and the M.S. and Ph.D. degrees in CS from the University of California at Los Angeles (UCLA), Los Angeles, CA, USA, in 2018 and 2022, respectively.

He has worked at Arm Research, Austin, TX, USA; IBM Research, New Delhi, India; Oracle, Bengaluru, India; and Teradata Labs, El Segundo, CA, USA. He is currently an Applied Scientist with Amazon, Seattle, WA, USA. To date, he has published more than 40 peer-reviewed conference papers, journal articles, and book chapters. His research explores robust machine learning and reinforcement learning system design for production scale applications, as well as privacy in the context of scalable data-driven machine learning for Internet-of-Things infrastructures.

**Mani Srivastava** (Fellow, IEEE) is currently on the Faculty at the University of California at Los Angeles (UCLA), Los Angeles, CA, USA, where he is with the Electrical and Computer Engineering (ECE) Department with a joint appointment in the Computer Science (CS) Department. He is a distinguished professor, the vice-chair of computer engineering, and an Amazon Scholar. His research is broadly in the area of human–cyber–physical and Internet-of-Things (IoT) systems that are learning-enabled, energy-efficient, secure, privacy-aware, and application-driven. It spans problems across the entire spectrum of applications, architectures, algorithms, and technologies in the context of systems and applications for mHealth, sustainable buildings, and smart built environments.

Dr. Srivastava is a Fellow of the Association for Computing Machinery (ACM). His research has been recognized by multiple best paper awards and the 10-Year Impact Award at the ACM International Symposium on Wearable Computing.